%% file: main.tex
\documentclass{article}
\PassOptionsToPackage{numbers,sort&compress}{natbib}
\usepackage[preprint]{neurips_2026}
\usepackage{microtype}
\usepackage[dvipsnames]{xcolor}
\usepackage{graphicx,amsmath,amssymb,booktabs}
\usepackage[format=plain,labelformat=simple,labelsep=period,font=small]{caption}
\usepackage[font=footnotesize,skip=3pt,subrefformat=parens]{subcaption}
\usepackage{float}
\usepackage[hyphens]{url}
\usepackage[shortlabels,inline]{enumitem}
\setlist[itemize]{noitemsep,leftmargin=*,topsep=0em}
\setlist[enumerate]{noitemsep,leftmargin=*,topsep=0em}
\usepackage[ruled,vlined,linesnumbered]{algorithm2e}
\input{preamble}
\definecolor{linkblue}{rgb}{0.21,0.49,0.74}
\usepackage[breaklinks,colorlinks,allcolors=linkblue]{hyperref}
\hypersetup{pdftitle={Continual Concept Personalization in Diffusion Models via Residual-Space Adaptation}}
\newcommand{\giulia}[1]{}
\newcommand{\enis}[1]{}
\newcommand{\yz}[1]{}
\title{Continual Concept Personalization in Diffusion Models via Residual-Space Adaptation}
\input{authors}
\date{}
\begin{document}
\maketitle
\input{sec/0_abstract}
\input{sec/1_intro}

\input{sec/2_related_work}

\input{sec/3_evaluation}
\input{sec/4_method}
\input{sec/5_continual_personalization_performance}

\input{sec/6_layer_probing_analysis}

\input{sec/8_conclusion}

\clearpage
{\small
\bibliographystyle{ieeenat_fullname}
\bibliography{main}
}
\clearpage
\appendix
\input{supp/1_metric_comparison}
\input{supp/2_stochastic_derivation}
\input{supp/3_implementation_details}
\input{supp/4_layer_mapping}
\input{supp/5_qualitative_examples}
\end{document}

%% file: preamble.tex
\usepackage{placeins}


%% file: authors.tex
\author{%
Yuzhi Liu\thanks{Corresponding to: \texttt{yuzliu@student.ethz.ch}} \quad
Enis Simsar \quad
Giulia Lanzillotta\quad
Thomas Hofmann\\[0.4em]
ETH Z\"urich%
}
\hypersetup{pdfauthor={Yuzhi LIU, Enis Simsar, Giulia Lanzillotta, Thomas Hofmann}}

%% file: sec/0_abstract.tex
\begin{abstract}

Text-to-image personalization aims to equip pretrained diffusion models with the ability to learn user-specific concepts from only a few reference images. Existing methods, however, typically train each concept independently or assume that all concepts are available simultaneously. In practical applications, concepts often arrive sequentially, requiring models to continually learn new subjects, objects, or styles while minimizing interference with previously acquired concepts. We address this continual concept personalization setting by extending residual-space adaptation to the stochastic adaptation process of pretrained diffusion models. Our work achieves the highest concept fidelity among competing continual personalization methods while maintaining near-zero forgetting. Beyond these empirical gains, the learned residual-space dynamics reveal where adaptation capacity is concentrated within the diffusion model. Restricting adaptation to the identified layers preserves nearly the same performance while reducing the number of trainable parameters by up to 72\%, demonstrating that residual-space dynamics can identify the layers that contribute most to continual concept personalization and directly guide more parameter-efficient adaptation. 

\end{abstract}

%% file: sec/1_intro.tex
\section{Introduction}
\label{sec:intro}


Recent text-to-image models such as Stable Diffusion~\cite{rombach2022high,podell2023sdxlimprovinglatentdiffusion} and FLUX~\cite{labs2025flux1kontextflowmatching} have enabled high-quality image synthesis from natural language descriptions, making generative models increasingly accessible to end users. Their flexibility has further motivated personalization methods that adapt pretrained models to user-specific concepts, subjects, objects, or artistic styles from only a few examples.

Personalization methods commonly rely on parameter-efficient fine-tuning techniques such as LoRA~\cite{hu2021loralowrankadaptationlarge} to adapt pretrained diffusion models to new concepts. Most existing approaches~\cite{gu2023mixofshowdecentralizedlowrankadaptation,po2024orthogonal,kumari2023multi,simsar2025loraclrcontrastiveadaptationcustomization} perform a separate fine-tuning procedure for every concept, and subsequently merge the resulting adapters into a unified personalized model. In many practical scenarios, however, concepts are acquired sequentially over time as new subjects, objects, or styles become available. Under this setting, whenever a new concept is introduced, 
the expensive merging procedure must be repeated to reconstruct the final personalized model. As the number of concepts grows, this repeated merging process becomes increasingly inefficient. This motivates continual concept personalization, where a single model is updated incrementally to incorporate new concepts while preserving previously acquired ones. 
The main obstacle to continual concept personalization is interference: learning new concepts can degrade performance on previously learned concepts. At the same time, the fixed model capacity must be used parsimoniously to preserve sufficient adaptation capacity for future concepts. 
Several continual learning methods \citep{farajtabarOrthogonalGradientDescent2020, saha2021gradient, liang2024inflora, barinkaprovable} achieve low levels of interference by constraining the parameter update to a pre-computed safe subspace. The most successful among these rely on layer-wise input statistics to characterize the safe subspace. This procedure cannot be directly extended to stochastic backbones, thus limiting its applicability in image generation tasks. 
In this work, we bridge this gap, extending residual-space adaptation to stochastic diffusion models for continual concept personalization. In particular, our starting point is an adapted version of DyRA~\cite{barinkaprovable} into diffusion models, a recent method which we choose for its elegant formulation and explicit theoretical guarantees.
We also introduce a novel systematic evaluation framework for continual concept personalization and show that the resulting residual-space dynamics reveal layerwise specialization, providing a practical tool for guiding parameter-efficient adaptation.

\paragraph{Contributions.} Our contributions are as follows:

\begin{itemize}
    \item We extend residual-space adaptation from deterministic language models to stochastic diffusion training, providing a new competitive option for continual concept personalization that mitigates interference while maintaining adaptivity at low cost.

    \item We introduce a systematic evaluation framework for continual personalization that captures concept fidelity, going beyond judging only image-generation quality.

    \item We show that residual-space dynamics can serve as an efficient probe for identifying the layers that contribute most to continual concept adaptation, providing insights into layerwise specialization and guiding more targeted parameter-efficient fine-tuning strategies.
\end{itemize}

%% file: sec/2_related_work.tex
\section{Related Work} \label{sec:related_work}

\subsection{Diffusion Personalization}

Text-to-image personalization has been widely studied for adapting pretrained diffusion models to user-specific identities, objects, or styles from only a few examples. Early approaches such as Textual Inversion~\cite{gal2022image} learn new concepts in the text embedding space while keeping the diffusion model fixed. DreamBooth~\cite{ruiz2023dreambooth} improves personalization quality by fine-tuning the diffusion model using a prior-preservation objective. Subsequent methods such as Custom Diffusion~\cite{kumari2023multi} restrict adaptation to selected cross-attention layers, improving parameter efficiency while supporting multi-concept customization. More recent personalization methods increasingly adopt parameter-efficient fine-tuning techniques such as LoRA. Mix-of-Show~\cite{gu2023mixofshowdecentralizedlowrankadaptation} introduces ED-LoRA together with structured concept tokens for multi-concept personalization, while LoRACLR~\cite{simsar2025loraclrcontrastiveadaptationcustomization} improves composition through contrastive alignment of independently trained adapters.

Another line of work reduces interference through the design of the adaptation space itself. Orthogonal Adaptation (OA)~\cite{po2024orthogonal} constructs approximately orthogonal LoRA subspaces, allowing independently trained adapters to be combined through simple summation with minimal interference. Although originally proposed in a non-continual setting, its orthogonal structure naturally lends itself to sequential adaptation. Overall, most existing personalization methods assume that all concepts are jointly available during training or are learned independently and combined afterward, and are therefore not explicitly designed for continual settings.

\subsection{Continual Adaptation}

Continual learning methods \citep{hadsell2020embracing} operate in a sequential training fashion, aiming to preserve knowledge and plasticity over time. Classical approaches rely on regularization, replay, or gradient projection mechanisms to mitigate catastrophic forgetting~\cite{wang2024comprehensive}.

Within diffusion adaptation, C-LoRA~\cite{smith2023continual} extends Custom Diffusion to a sequential setting by introducing regularization to preserve earlier concepts. Orthogonal Adaptation~\cite{po2024orthogonal} can also be naturally extended to continual settings through additive composition of orthogonal adapters.

Several methods constrain the parameter update to a safe subspace based on the layer-wise input statistics. Among these, 
Gradient Projection Memory (GPM)~\cite{saha2021gradient} estimates layer-wise feature covariance from previous tasks and constrains the new updates to its low-energy subspace.
The more recent InfLoRA~\cite{liang2024inflora} applies the same strategy to LoRA-based fine-tuning, where each task is learned within a fixed low-rank budget. 
Building upon InfLoRA, Barinka~et~al.~\cite{barinkaprovable} propose a dynamic rank allocation method with simplified updates which allows for significant capacity savings. In addition to providing theoretical guarantees for mitigating forgetting, the method demonstrates strong empirical performance on challenging continual language model adaptation tasks.

Although effective, these approaches have been studied primarily in deterministic architectures. Whether residual-space methods remain effective in stochastic diffusion training, where layer statistics are influenced by sampled noise, remains unexplored.

\subsection{Evaluation of Personalized Generative Models} \label{sec:related_work_eval}

Evaluating personalized generative models typically involves measuring text alignment, image similarity, and identity preservation. Text alignment is commonly assessed using CLIP-based~\cite{radford2021learning} text-image similarity, which measures whether generated images are semantically consistent with their prompts. Image alignment metrics compare generated images with reference examples, often using CLIP image embeddings. Several works additionally evaluate identity preservation through specialized recognition models such as ArcFace~\cite{deng2019arcface}. However, these metrics are mainly designed for human faces and do not generalize to arbitrary objects.

Beyond specialized identity metrics, a variety of pretrained embedding models have been used to assess perceptual similarity. CLIP~\cite{radford2021learning} is trained on large-scale image--text pairs and provides representations that emphasize global semantic correspondence between images and language. In contrast, DINO-based models~\cite{caron2021emerging,oquab2023dinov2,simeoni2025dinov3} learn self-supervised visual representations without text supervision and preserve fine-grained visual structure and local semantic information. DreamSim~\cite{fu2023dreamsim} combines multiple pretrained visual embeddings together with human similarity annotations, leveraging both global semantic understanding and fine-grained visual details to produce perceptual representations that closely align with human judgments.

Overall, most existing evaluation protocols are designed for static personalization settings and do not measure identity preservation or forgetting over time.

%% file: sec/3_evaluation.tex
\section{Evaluating Continual Performance}
\enis{suggestion: Measuring Forgetting in Continual Personalization}
Evaluating continual text-to-image personalization requires measuring not only the quality of generated images but also how well previously learned concepts are preserved throughout sequential adaptation. Unlike discriminative tasks, where forgetting can be directly quantified through accuracy, there is no canonical notion of forgetting for generative models.
As discussed in Section~\ref{sec:related_work_eval}, existing evaluation protocols are primarily designed for static personalization settings and focus on aggregate measures of text or image alignment. In order to assess the degradation of the concept identity over time, it is important to have metrics that distinguish concept fidelity from general image quality. 
To address this problem, we first introduce a concept-level fidelity measure that quantifies how faithfully a personalized concept is preserved. We then derive continual evaluation metrics directly from this quantity.

\subsection{Concept Fidelity}

Our objective is to evaluate the personalized concept itself rather than the similarity between individual images. A generated image may differ substantially from its reference in pose, background, or artistic style while still faithfully depicting the same concept. We therefore introduce \emph{concept fidelity}, which measures similarity at the concept level rather than the image level.

To quantify concept fidelity, we first measure similarity between individual generated and reference images in a learned embedding space. Given an embedding function $F(\cdot)$, we define the similarity between two images $i_1$ and $i_2$ as
\begin{equation} \label{eq:sim}
\mathrm{sim}(i_1,i_2)
=
\cos\left(F(i_1),F(i_2)\right).
\end{equation}

While image similarity is defined between individual image pairs, the fidelity of a personalized concept should be assessed from a set of generated images rather than a single sample. We therefore generalize the image-level similarity measure by aggregating pairwise similarities across the generated and reference image sets.

Let $G=\{g_1,\dots,g_m\}$ denote the generated images corresponding to a personalized concept and let $I=\{i_1,\dots,i_n\}$ denote the associated reference images. The concept fidelity score is defined as
\begin{equation}
S(G,I)
=
\frac{1}{|G||I|}
\sum_{g\in G}
\sum_{i\in I}
\mathrm{sim}(g,i).
\end{equation}

This concept fidelity score extends image-level similarity to a set of generated images, providing a single measure of how well a model preserves a personalized concept.

\begin{figure}[!htbp]
	\centering
	\includegraphics[width=0.72\linewidth,height=0.46\textheight,keepaspectratio]{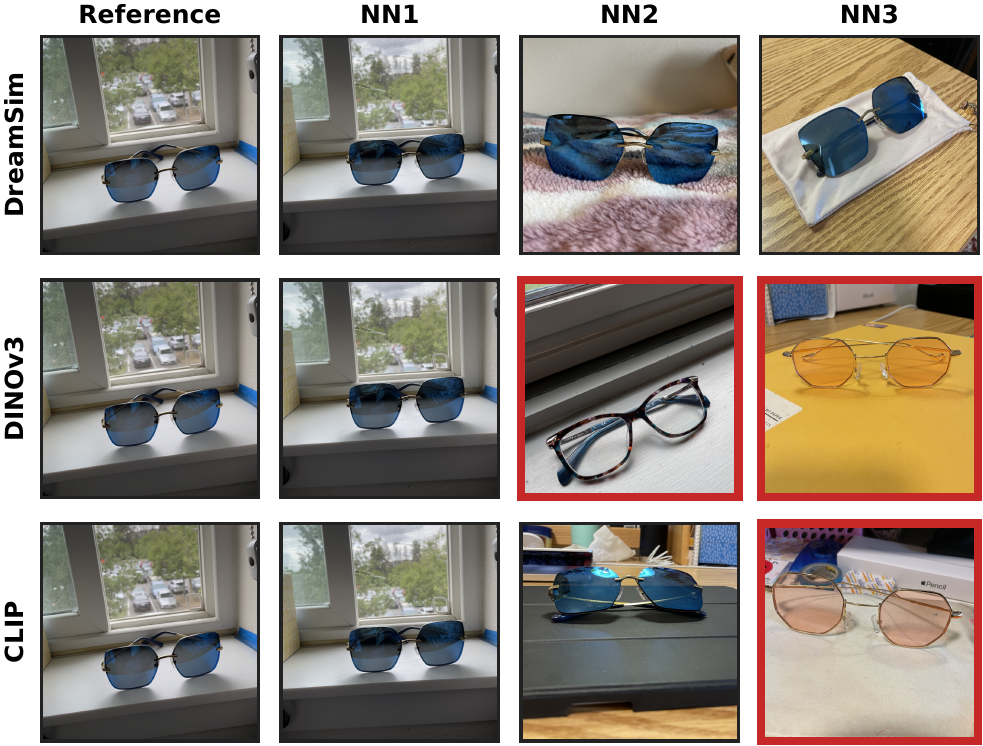}
	\caption{Nearest-neighbor retrieval for one reference image across embedding spaces. Red frames indicate retrieved images that belong to a different concept than the query.}
	\label{fig:embedding_nn_comparison}
\end{figure}

We consider several pretrained embedding models, including CLIP~\cite{radford2021learning}, DINO~\cite{simeoni2025dinov3}, and DreamSim~\cite{fu2023dreamsim}, as candidates for computing concept fidelity. Figure~\ref{fig:embedding_nn_comparison} provides a qualitative example of nearest-neighbor retrieval using the distance ($1-\mathrm{sim}$), where $\mathrm{sim}$ denotes the similarity score defined in Eq.~\eqref{eq:sim}. DreamSim retrieves images belonging to the same personalized concept despite variations in pose and background, whereas CLIP and DINO retrieve visually similar images from different concepts. Our quantitative comparison in Appendix~\ref{sup:metric_comparison} further supports that image similarity in the DreamSim embedding space best reflects concept-level similarity. We therefore adopt DreamSim as the default embedding model throughout the paper.

\subsection{Final Fidelity and Forgetting Metrics}
Using the concept fidelity score, we now define continual evaluation metrics that summarize model performance throughout sequential adaptation.

Let $N$ denote the total number of concepts. For concept $k$, let $G_k^{(T)}$ denote the generated images after all concepts have been learned. 

\paragraph{Final concept fidelity.} We define the final concept fidelity as:
\begin{equation}
Q
=
\frac{1}{N}
\sum_{k=1}^{N}
S\left(G_k^{(T)},I_k\right),
\end{equation}
which characterizes the average concept fidelity retained at the end of continual training.

\paragraph{Forgetting dynamics and final forgetting.}
To investigate the dynamics of forgetting throughout training, let $t_k$ denote the training stage immediately after learning concept $k$. We define the average forgetting trajectory at training stage $t\ge t_k$ as 
\begin{equation} F(t) = \frac{1}{n_t} \sum_{k=1}^{n_t} \left[ S\left(G_k^{(t_k)},I_k\right) - S\left(G_k^{(t)},I_k\right) \right], \end{equation} where $n_t$ is the number of concepts that have been learned by stage $t$. The trajectory measures the average loss of concept fidelity as new concepts are acquired.

The \textbf{final forgetting} is defined as
\begin{equation} F = F(T), \end{equation} 
which measures the average loss of concept fidelity after continual training. 

Together, $Q$ summarizes the model's average concept fidelity after continual training, while $F(t)$ characterizes the forgetting dynamics throughout sequential adaptation.

%% file: sec/4_method.tex
\section{Residual-Space Adaptation in  Diffusion}

We build upon the residual-space adaptation framework DyRA~\cite{barinkaprovable}, originally designed for LLMs, which constructs layer-wise adaptation subspaces from the input representations. Applying this framework to diffusion models is nontrivial, as these representations depend not only on the input concept but also on the randomly sampled diffusion noise and denoising timestep. We therefore first review the residual-space adaptation procedure and then develop a stochastic interpretation that justifies applying the framework to diffusion models and motivates our implementation.


\subsection{Background: DyRA}

We consider a sequence of concept personalization tasks
$\{\mathcal D_t\}_{t=1}^T$, where each task introduces a new concept $t$ and
training is performed sequentially. Following
\cite{barinkaprovable}, each adapted layer $l$ is parameterized by a LoRA
update
\begin{equation}
\Delta W_t^{(l)} = A_t^{(l)}B_t^{(l)},
\end{equation}
where $B_t^{(l)}$ is initialized before training and kept fixed, while
only $A_t^{(l)}$ is optimized. The accumulated model after task $t$ is
\begin{equation}
W_t^{(l)} = W^{(l)} + \sum_{i=1}^{t}\Delta W_i^{(l)}.
\end{equation}

The key idea of DyRA is to restrict learning to feature directions that are informative for the current task while avoiding directions already captured by previous tasks. For each adapted layer, the method applies a spectral decomposition of the input features matrix from all tasks, and identifies therein directions unique to the current task, i.e. \emph{the residual space} leftover from previous tasks. Since different layers exhibit different residual-space spectra, the number of informative directions also varies, motivating adaptive rank allocation instead of a fixed LoRA rank.

Concretely, the procedure for each adapted layer consists of three
steps:
\begin{enumerate}
\item Estimate the input subspace most representative of previous tasks, named $\mathcal M_{t-1}^{(l)}$ .

\item Project the current task input space
      $\mathcal H_t^{(l)}$ onto the orthogonal complement
      of $\mathcal M_{t-1}^{(l)}$ to obtain the residual space.

\item Initialize the LoRA down-projection matrix
      $B_t^{(l)}$ in the residual space, picking only
      directions with singular values above a given threshold $\tau$. During training,
      $B_t^{(l)}$ remains fixed and only the corresponding
      up-projection matrix $A_t^{(l)}$ is optimized.
\end{enumerate}
Importantly,  it is the threshold which implicitly determines the layer-wise rank. This is the key to flexible rank allocation.

\subsection{A Stochastic Formulation for Diffusion}

A key step of residual-space adaptation is constructing the residual space from layer input representations collected before training each task. For deterministic backbones, this procedure is well defined since each training sample corresponds to a unique layer input. Diffusion models, however, introduce an additional source of stochasticity through the sampled diffusion noise and denoising timestep, so the layer input of a training sample is no longer unique. This raises the central question of this section: \emph{how should the layer input space
$\mathcal H_t^{(l)}$ be estimated for residual-space construction?} 

Start by interpreting the layer input as the random variable
\begin{equation}
x_t^{(l)} = x_t^{(l)}(s,\epsilon),
\end{equation}

where $\epsilon\sim\mathcal N(0,I)$ denotes the sampled diffusion noise and $s$ denotes the sampled denoising timestep.

Extending the deterministic interference measure in~\cite{barinkaprovable}, $\|B_j^{(l)}x_i^{(l)}\|^2$ to its expectation,
consider the following stochastic interpretation of residual-space adaptation: projection directions should maximize the expected adaptation ability of the current task while suppressing the expected interference with previously learned tasks,
\begin{equation}
\begin{aligned}
\max_{B_j^{(l)}}\quad &
\mathbb E_{s,\epsilon}
\left[
\left\|
B_j^{(l)}
x_j^{(l)}
\right\|^2
\right] \\
\text{s.t.}\quad &
\mathbb E_{s,\epsilon}
\left[
\left\|
B_j^{(l)}
x_i^{(l)}
\right\|^2
\right]
\le \delta,\quad i<j.
\end{aligned}
\end{equation}

From this interpretation the answer to our question directly follows. Since the optimization objective is defined over the training distribution of diffusion noise and denoising timesteps, the collected layer inputs should provide an empirical estimate of the same distribution. Therefore, layer inputs are collected using exactly the same sampling strategy as diffusion training, namely Gaussian noise with uniformly sampled denoising timesteps. This simple step extends residual-space adaptation from deterministic feature representations to stochastic diffusion training by aligning the feature collection procedure with the stochastic optimization objective. Detailed implementation and derivation are provided in Appendix~\ref{sup:stochastic_derivation}.

%% file: sec/5_continual_personalization_performance.tex
\section{Stable Continual Personalization}
\label{sec:performance}

This section evaluates whether residual-space adaptation can be successfully extended to continual diffusion personalization. We first assess its ability to preserve concept fidelity while mitigating forgetting throughout continual learning, and then examine scalability and capacity preservation.

\subsection{Experimental Setup}
We conduct all experiments on Concept101~\cite{kumari2023multi}, a benchmark containing 101 concepts spanning 17 semantic classes, including visually similar concepts, making it particularly suitable for studying interference and identity preservation in continual learning. All concepts are learned sequentially in a rehearsal-free setting, where only the current concept data are available during training. Unless otherwise stated, all methods follow the same randomly shuffled task ordering. Throughout the experiments, we denote DyRA configured with the sigular value threshold $\tau$ as DyRA-$\tau$. We compare \textbf{DyRA} with $\tau\in\{0.5,1.0\}$ against three continual LoRA baselines: \textbf{Trainable-AB}, which optimizes both LoRA projection matrices $A,B$ and corresponds to a continual variant of Mix-of-Show~\cite{gu2023mixofshowdecentralizedlowrankadaptation}; \textbf{Rank-Fixed-B}, which optimizes only the LoRA up-projection $A$ while keeping the randomly initialized down-projection $B$ fixed; and \textbf{OA}, the continual version of Orthogonal Adaptation~\cite{po2024orthogonal}. Unless otherwise specified, all methods adapt the linear projections in the transformer blocks of both the text encoder and the UNet, and all fixed-rank methods use rank $r=4$. All experiments are conducted on Stable Diffusion v1.5~\cite{rombach2022high}. Models are periodically evaluated throughout continual training. For each concept, images are generated using only the learned concept token as the prompt, without additional textual descriptions, thereby isolating the learned concept representation and enabling a direct comparison of concept fidelity across methods. At the end of training, we further validate the model using diverse prompt templates. Additional implementation details and training hyperparameters are provided in Appendix~\ref{sup:implementation_details}.

\subsection{Concept Fidelity and Forgetting}

\paragraph{Quantitative Results.}

Table~\ref{tab:cl_dreamsim_avg} summarizes the final concept fidelity ($Q$) and forgetting ($F$). Trainable-AB achieves competitive concept fidelity but exhibits the largest forgetting, highlighting the trade-off between adapting to new concepts and preserving previously learned ones. In contrast, Rank-Fixed-B and OA effectively suppress forgetting but achieve slightly lower concept fidelity. DyRA maintains negligible forgetting while achieving the highest concept fidelity. These results demonstrate that residual-space adaptation extends effectively to the stochastic training process of diffusion models.

\begin{table}[!htbp]
\centering
\small
\setlength{\tabcolsep}{5pt}
\renewcommand{\arraystretch}{0.92}
\begin{tabular}{lcc}
\toprule
Method & $Q\,\uparrow$ & $F\,\downarrow$ \\
\midrule
Trainable-AB      & 0.6392 $\pm$ 0.0015 & 0.0368 $\pm$ 0.0014 \\
Rank-Fixed-B      & 0.6442 $\pm$ 0.0029 & \textbf{$-$0.0033 $\pm$ 0.0020} \\
OA                & 0.6435 $\pm$ 0.0031 & $-$0.0023 $\pm$ 0.0003 \\
DyRA-0.5        & 0.6454 $\pm$ 0.0022 & 0.0002 $\pm$ 0.0001 \\
DyRA-1.0        & \textbf{0.6497 $\pm$ 0.0008} & $-$0.0001 $\pm$ 0.0007 \\
\bottomrule
\end{tabular}

\caption{Concept fidelity ($Q$) and forgetting ($F$) scores. Values are mean $\pm$ SD over three seeds.}

\label{tab:cl_dreamsim_avg}
\end{table}

\begin{figure}[!htbp]
	\centering
	\begin{subfigure}{0.58\linewidth}
		\centering
		\includegraphics[width=\linewidth]{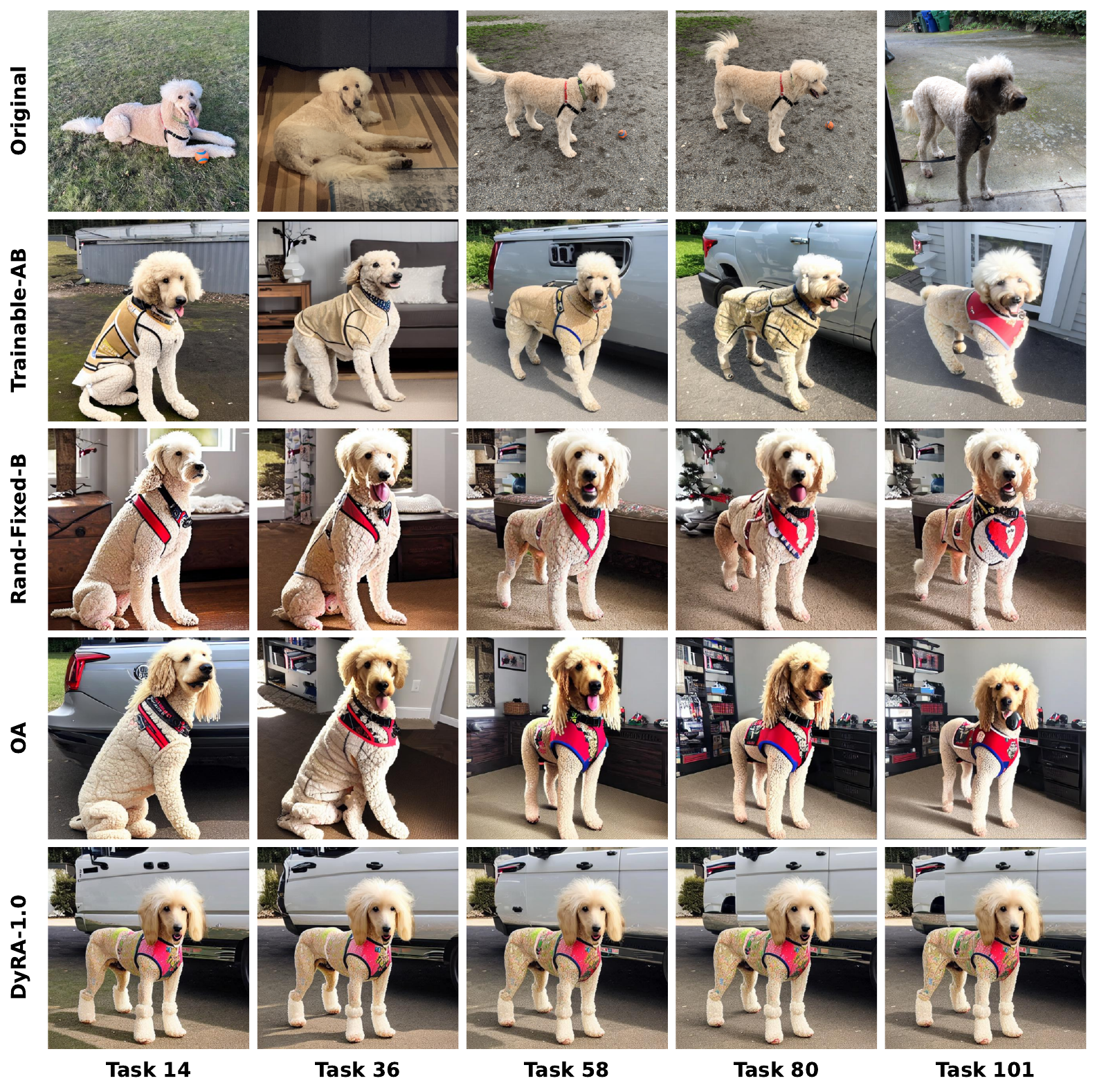}
		\caption{Continual generation across the task sequence. Row~1 shows reference images; rows~2--5 show Trainable-AB, Rank-Fixed-B, OA, and DyRA-$1.0$.}
		\label{fig:cg_petdog2}
	\end{subfigure}
	\par\medskip
	\begin{subfigure}{0.58\linewidth}
		\centering
		\includegraphics[width=\linewidth]{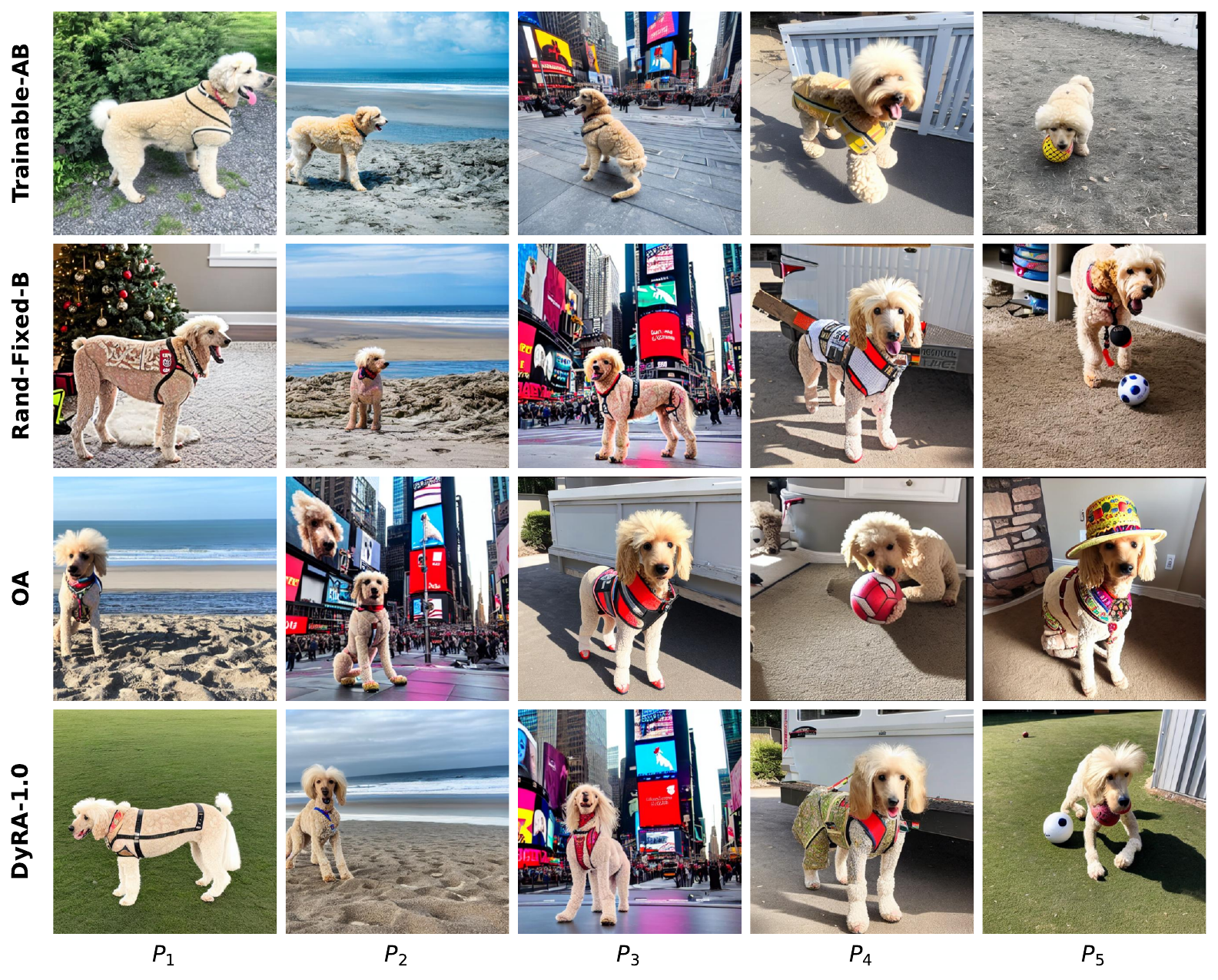}
		\caption{Final generations under diverse prompt templates. Rows show Trainable-AB, Rank-Fixed-B, OA, and DyRA-$1.0$.}
		\label{fig:val_petdog2}
	\end{subfigure}
	\caption{Qualitative results for concept~14 (\textit{pet\_dog2}).}
\end{figure}

\paragraph{Qualitative Results.}
The quantitative improvements are also reflected in the generated images. Figure~\ref{fig:cg_petdog2} tracks concept \textit{pet\_dog2}, introduced at task~14, from the moment it is learned to the final model. As additional concepts are introduced, Trainable-AB exhibits the largest appearance drift, Rank-Fixed-B and OA preserve the concept with only minor degradation, and DyRA remains nearly unchanged. Figure~\ref{fig:val_petdog2} shows that DyRA maintains high concept fidelity under diverse prompt templates, demonstrating generalization beyond the token-only prompts used during continual evaluation.

\paragraph{Forgetting dynamics.}

\begin{figure}[!htbp]
	\centering
	\includegraphics[width=0.78\linewidth,height=0.42\textheight,keepaspectratio]{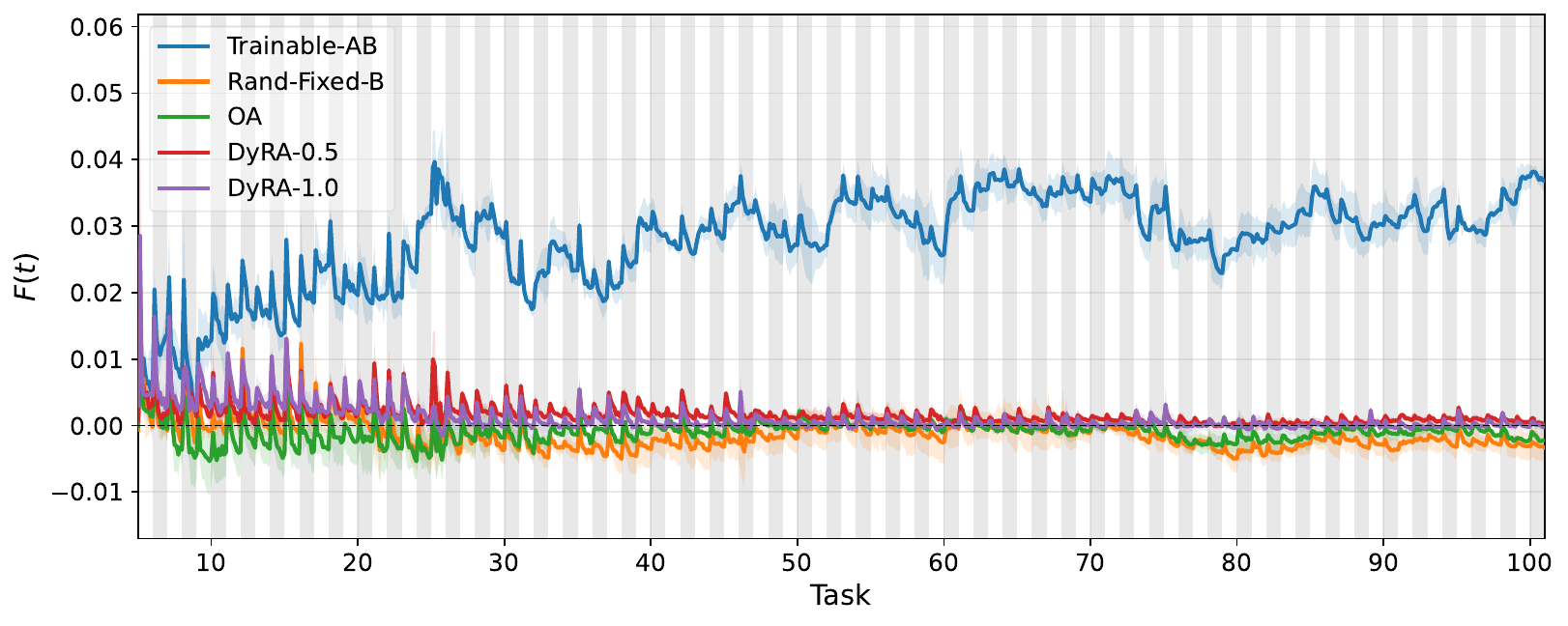}
	\caption{Forgetting trajectory across continual training. Higher values indicate more forgetting.}
	\label{fig:forgetting_traj}
\end{figure}

\begin{figure}[!htbp]
	\centering
	\includegraphics[width=0.76\linewidth,height=0.44\textheight,keepaspectratio]{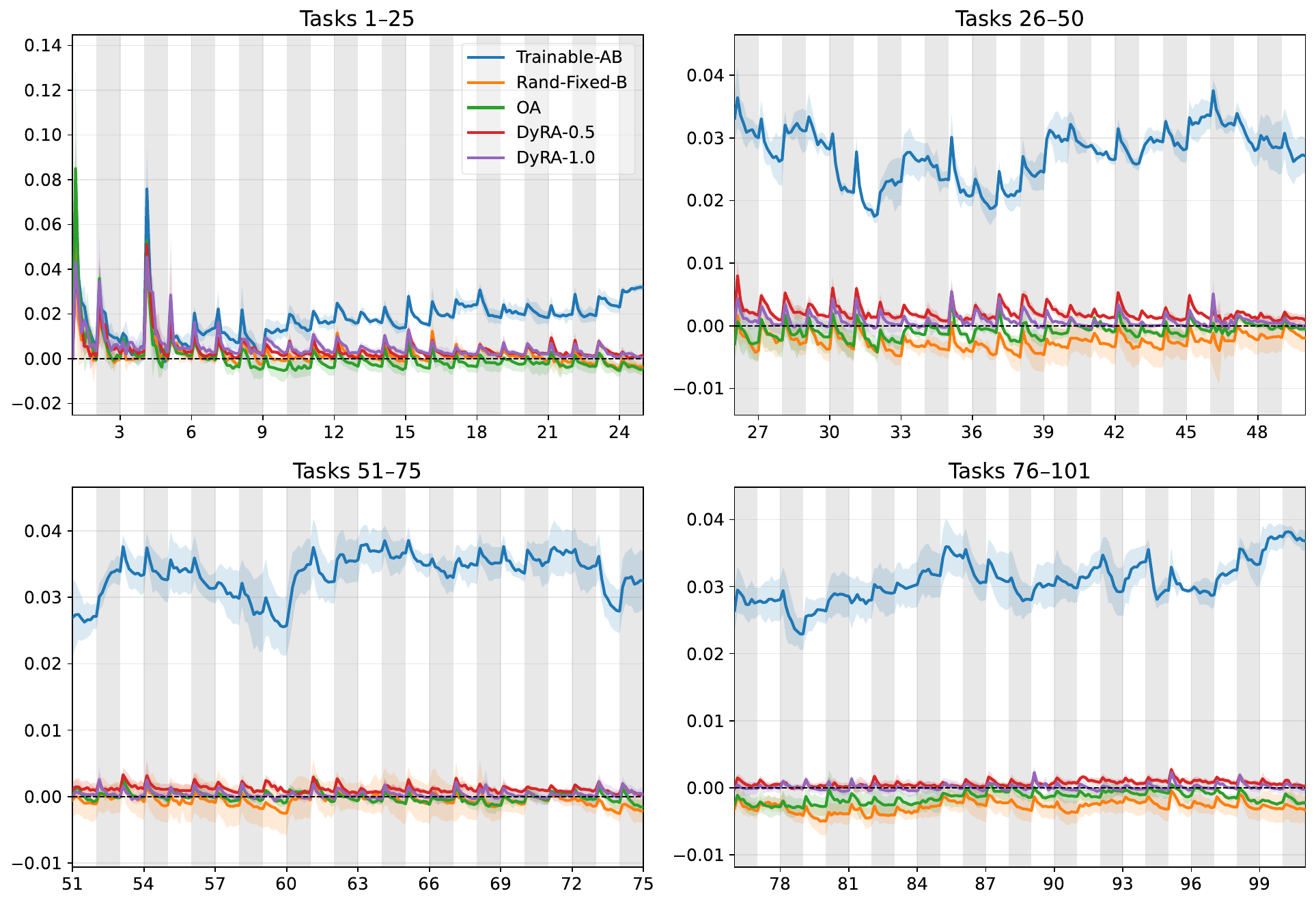}
	\caption{Zoomed views of the forgetting trajectory over four task intervals.}
	\label{fig:forgetting_traj_zoomed}
\end{figure}

While Table~\ref{tab:cl_dreamsim_avg} summarizes the final forgetting score, Figure~\ref{fig:forgetting_traj} reveals how forgetting evolves throughout continual learning. Trainable-AB oscillates around substantially higher forgetting values, indicating persistent forgetting throughout the training process. In contrast, Rank-Fixed-B, OA, and DyRA all fluctuate closely around zero, demonstrating that their low forgetting is maintained consistently over time rather than occurring only at the final evaluation. Figure~\ref{fig:forgetting_traj_zoomed} provides a zoomed-in view, enabling a clearer comparison of the forgetting dynamics among Rank-Fixed-B, OA, and DyRA.

Overall, the quantitative metrics, qualitative generations, and forgetting trajectories consistently demonstrate that DyRA preserves high concept fidelity while effectively mitigating catastrophic forgetting.

\subsection{Dynamic Rank Improves Parameter Efficiency}
The performance gains of DyRA stem not only from restricting adaptation to the residual space, but also from allocating adaptation capacity according to the residual-space structure of each layer. We therefore compare DyRA with the fixed-rank OA under different trainable parameter budgets to evaluate the effectiveness of dynamic rank allocation.

\begin{figure}[!htbp]
	\centering
	\includegraphics[width=0.66\linewidth,height=0.42\textheight,keepaspectratio]{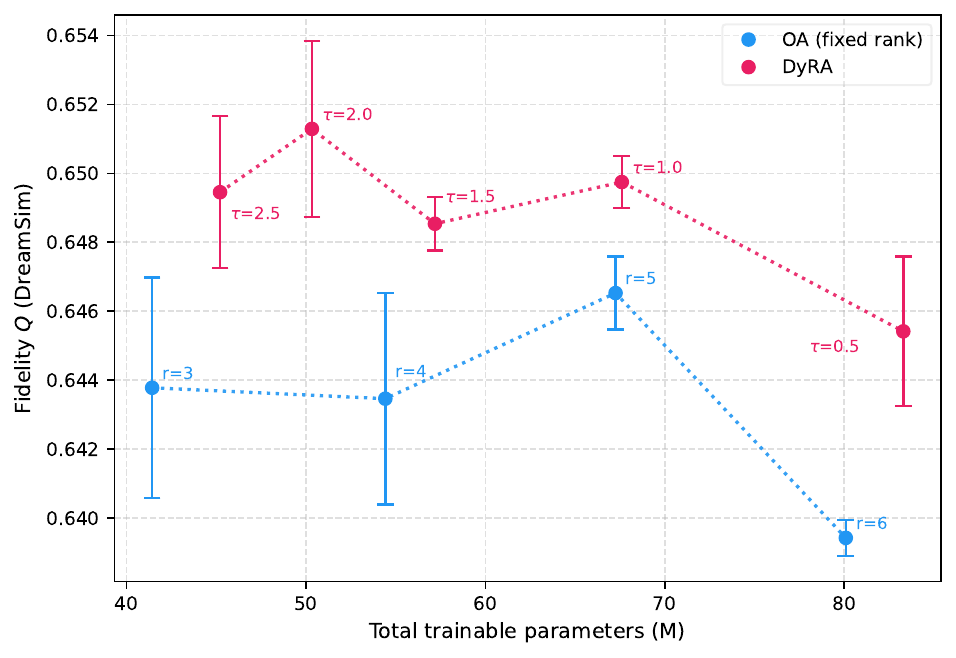}
	\caption{Parameter-efficiency frontier comparing concept fidelity and trainable parameter count for OA and DyRA variants. Error bars indicate one standard deviation over three random seeds.}
	\label{fig:efficiency_frontier}
\end{figure}

Figure~\ref{fig:efficiency_frontier} compares concept fidelity against the number of trainable parameters for OA with different fixed ranks and DyRA with different singular-value thresholds. Each point corresponds to a different parameter budget. Across the entire range of budgets, DyRA consistently achieves higher concept fidelity than OA with comparable numbers of trainable parameters, demonstrating a superior performance--parameter trade-off.

Notably, even under high singular-value thresholds, where DyRA has substantially fewer trainable parameters, it outperforms OA with considerably larger parameter budgets. These results indicate that allocating rank adaptively according to the residual-space structure is substantially more parameter-efficient than assigning a uniform rank to every layer, allowing DyRA to maintain strong concept fidelity under a reduced parameter budget.

%% file: sec/6_layer_probing_analysis.tex
\section{Layer-wise Specialization}
Section~\ref{sec:performance} demonstrated that DyRA achieves strong continual personalization performance while improving parameter efficiency through adaptive rank allocation. Beyond these empirical gains, the adaptive ranks themselves provide a signal of where adaptation capacity is consumed during continual learning, since additional rank is allocated only when new concepts introduce directions outside the accumulated input subspace. This section analyzes these rank dynamics to understand how adaptation capacity is distributed across the layers and how this information can inform future parameter-efficient adaptation.

\subsection{Adaptation Capacity Concentration} \label{sec:6-1}
\begin{figure}[!htbp]
	\centering
	\includegraphics[width=0.86\linewidth,height=0.34\textheight,keepaspectratio]{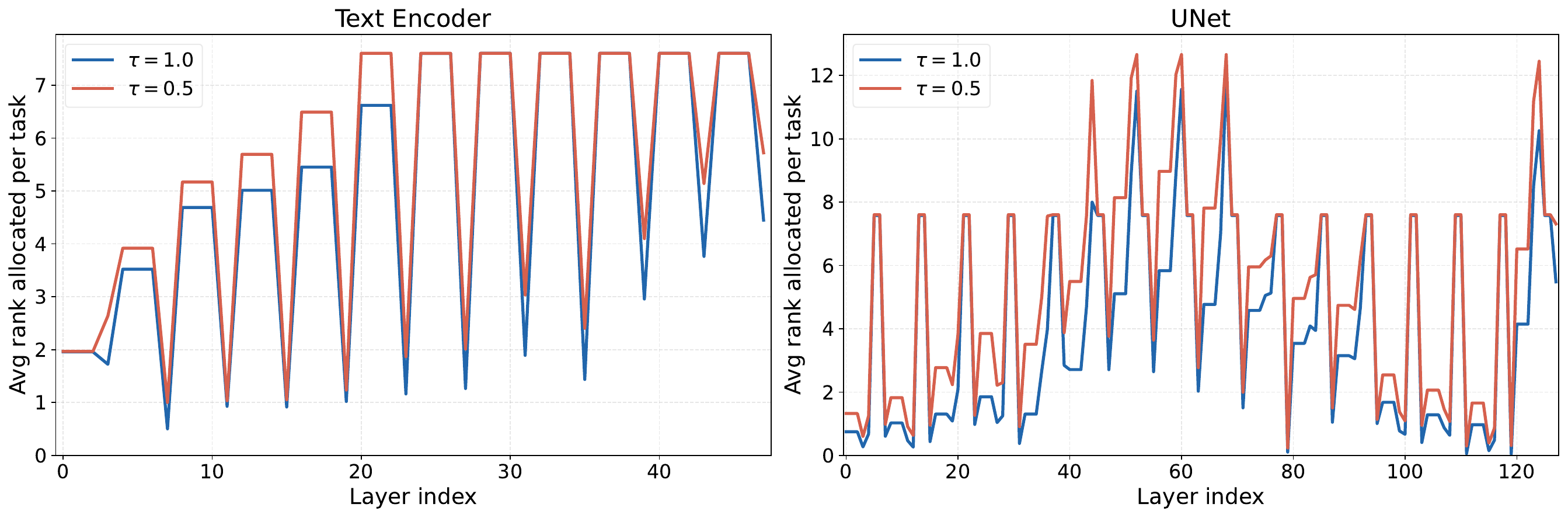}
	\caption{Average allocated rank per adapted layer over 101 concepts.}
	\label{fig:avg_rank_per_layer}
\end{figure}

\enis{suggestion -- Adaptation Concentrates in Cross-Attention K,V}
We first analyze how DyRA distributes adaptation capacity across the adapted layers by averaging the allocated ranks over all 101 concepts. Figure~\ref{fig:avg_rank_per_layer} illustrates the average rank allocated to each layer. The x-axis indexes the adapted layers in architectural order, and the corresponding module names are provided in Appendix~\ref{sup:layer_mapping}.


Rather than allocating the same rank to every layer, DyRA exhibits clear layer-dependent allocation patterns that closely follow the repeated transformer architecture. In the text encoder, the allocated ranks exhibit a four-layer periodicity that exactly mirrors the repeated key(K), value(V), query(Q), and output projection layers within each transformer block. A similar correspondence is observed in the UNet, where every group of eight adapted layers corresponds to the self-attention and cross-attention projections within a transformer block. Within every block, the two dominant rank peaks consistently coincide with the cross-attention key(K) and value(V) projection layers, which repeatedly receive substantially larger ranks than all other projections.

The repeated concentration of residual rank in the UNet cross-attention K,V projections indicates that these layers consistently consume more adaptation capacity than other adapted modules during continual concept personalization. This observation motivates a closer examination of these layers in the following analysis.

\subsection{Early Capacity Saturation} \label{sec:6-2}

The UNet cross-attention K,V projections receive the largest residual ranks. We next examine whether exhausting the residual capacity of these layers is associated with degraded adaptation to subsequently learned concepts.

\begin{figure}[!htbp]
	\centering
	\includegraphics[width=0.86\linewidth,height=0.34\textheight,keepaspectratio]{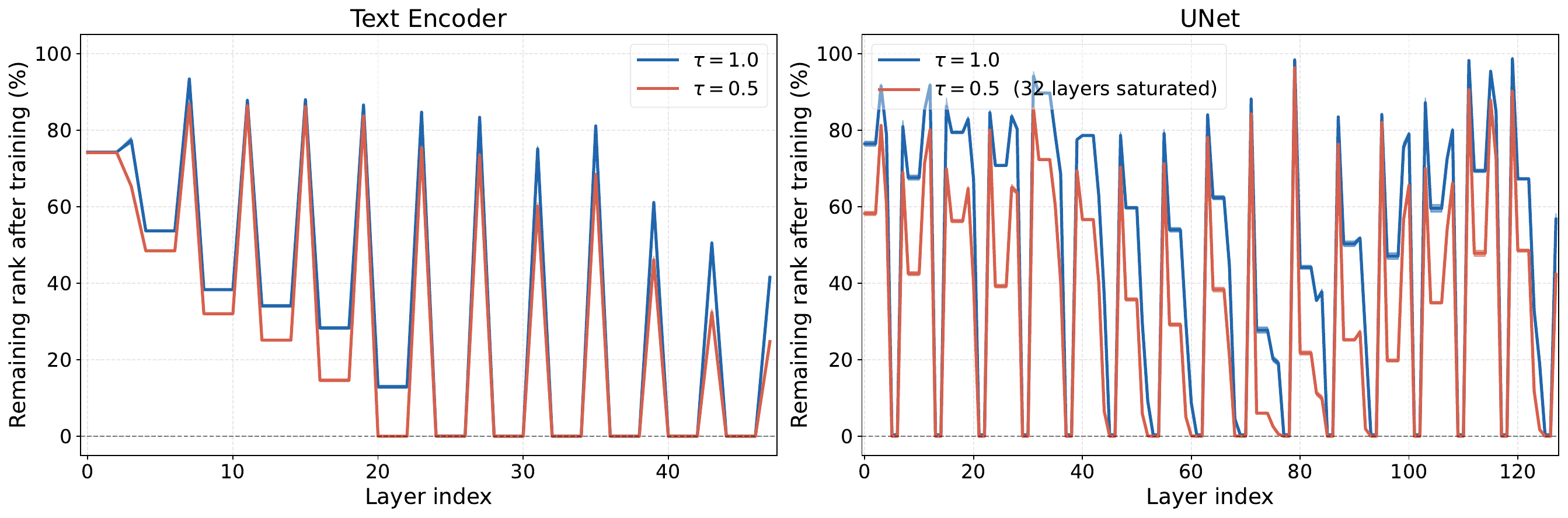}
	\caption{Remaining rank capacity after 101 tasks.}
	\label{fig:remaining_rank_final}
\end{figure}

\begin{figure}[!htbp]
	\centering
	\includegraphics[width=0.86\linewidth,height=0.34\textheight,keepaspectratio]{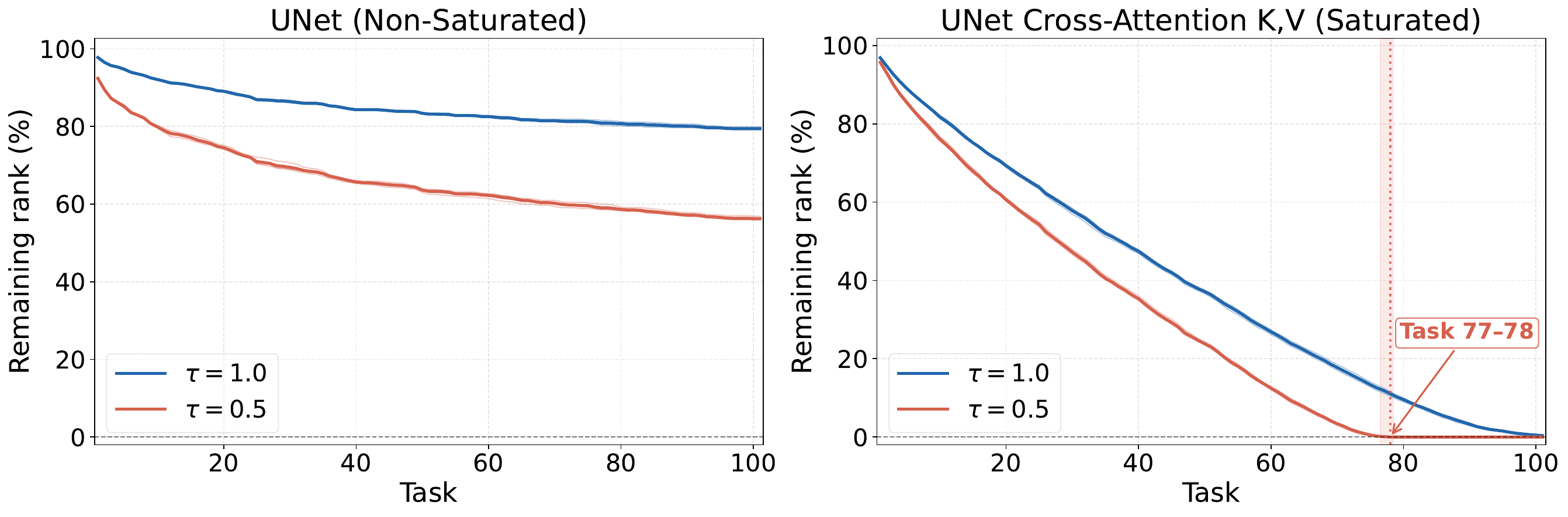}
	\caption{Representative remaining-rank trajectories of UNet layers: a non-saturated layer (left) and a cross-attention K,V layer that saturates under DyRA-$0.5$ (right).}
	\label{fig:remaining_rank_traj}
\end{figure}

Figure~\ref{fig:remaining_rank_final} summarizes the remaining rank of every adapted layer after learning all 101 concepts. Under DyRA-$0.5$, all 32 UNet cross-attention K,V projections become fully saturated, whereas no UNet layer saturates under DyRA-$1.0$. Figure~\ref{fig:remaining_rank_traj} further illustrates representative remaining-rank trajectories. While non-saturated layers retain sufficient residual capacity throughout training, representative cross-attention K,V layers under DyRA-$0.5$ steadily exhaust their available rank and become saturated around tasks 77--78. In contrast, the corresponding layers under DyRA-$1.0$ retain residual capacity throughout the entire task sequence.

To assess the effect of this capacity exhaustion, Table~\ref{tab:last20_fidelity} compares the final concept fidelity with the average fidelity of the final 20 learned concepts. DyRA-$1.0$ maintains essentially unchanged performance on later concepts, whereas DyRA-$0.5$ exhibits a clear late-task fidelity degradation after the cross-attention K,V layers become saturated. Although this evidence is correlational, the consistent agreement between the rank allocation, capacity saturation, and late-task performance suggests that preserving residual capacity in the UNet cross-attention K,V projections is important for sustaining continual concept adaptation over long task sequences.

\begin{table}[!htbp]
\centering
\small
\setlength{\tabcolsep}{4pt}
\begin{tabular}{lccc}
\toprule
Method & Final $Q\,\uparrow$ & Last-20 $Q\,\uparrow$ & Gap $\downarrow$\\
\midrule
DyRA-$0.5$ & 0.6454 & 0.6340 & 0.0114\\
DyRA-$1.0$ & 0.6497 & 0.6501 & -0.0004\\
\bottomrule
\end{tabular}
\caption{Final concept fidelity compared with the average fidelity on the final 20 learned concepts. The reported score are averaged over three seeds. Standard deviations are below 0.007 in all cases.}
\label{tab:last20_fidelity}
\end{table}

\subsection{Layer-Guided Parameter-Efficient Adaptation}\label{sec:6-3}

The analyses in Sections~\ref{sec:6-1} and~\ref{sec:6-2} consistently identify the UNet cross-attention K,V projections as the layers that require the largest residual adaptation subspaces throughout continual learning. This suggests that these layers differentiate newly learned concepts through distinct adaptation directions. If this interpretation is correct, restricting adaptation to the identified layers would be expected to preserve most of the continual personalization performance while substantially reducing the number of trainable parameters.

Table~\ref{tab:kv_ablation} supports this hypothesis. Restricting adaptation to only the UNet K,V projections reduces the trainable parameter count by approximately 60\% (67.56M $\rightarrow$ 27.12M) while achieving virtually identical concept fidelity ($Q=0.6496$ vs.\ $0.6497$) and comparable forgetting. Restricting adaptation further to only the UNet cross-attention K,V projections reduces the trainable parameter count by approximately 72\% (67.56M $\rightarrow$ 18.87M), yet still preserves nearly all continual personalization performance in practice.

Beyond validating the proposed interpretation, these results also provide a structural explanation for previous personalization methods such as Custom Diffusion~\cite{kumari2023multi} and C-LoRA~\cite{smith2023continual}, which restrict adaptation to cross-attention K,V projections based primarily on empirical observations of large weight updates during training. In contrast, residual-space dynamics explain the same design choice from the perspective of feature-space utilization, providing a principled interpretation of why these layers are particularly effective adaptation targets.

More broadly, residual-space dynamics provide a principled way to identify where adaptation capacity is most effectively utilized. Rather than relying on empirical heuristics or exhaustive layerwise search, the proposed analysis directly guides targeted adaptation, enabling substantially more parameter-efficient continual personalization while preserving nearly the same overall performance.

\begin{table}[!htbp]
\centering

\begin{tabular}{lccc}
\toprule
Method & $Q\,\uparrow$ & $F\,\downarrow$ & Params (M) \\
\midrule
Full DyRA& 0.6497 $\pm$ 0.0008 & $-$0.0001 $\pm$ 0.0007 & $67.56 \pm 0.06$ \\
UNet K,V & 0.6496 $\pm$ 0.0013 & $-$0.0011 $\pm$ 0.0003 & $27.12 \pm 0.06$ \\
UNet cross-attention K,V& 0.6479 $\pm$ 0.0004 & $-$0.0008 $\pm$ 0.0002 & $18.87 \pm 0.00$ \\
\bottomrule
\end{tabular}
\caption{Performance of targeted adaptation strategies guided by residual-space dynamics. Values are mean $\pm$ SD over three seeds.}
\label{tab:kv_ablation}
\end{table}

%% file: sec/8_conclusion.tex
\section{Conclusion}

\paragraph{Limitations and Future Directions.}
Although DyRA achieves strong continual personalization performance, several limitations remain. First, the adaptation capacity of residual-space methods is fundamentally bounded by the width of individual layers. As continual learning progresses, layers with consistently high adaptation demand may eventually exhaust their available residual subspaces, limiting scalability to arbitrarily long task sequences. Second, the method is implemented only on Stable Diffusion v1.5 and Concept101. Nevertheless, the residual-space analysis developed in this work is considerably more general than the specific adaptation algorithm. Since it relies only on layer input statistics and residual-space dynamics, it can potentially be applied to other diffusion backbones, transformer-based generative models, or broader neural architectures to identify the layers that contribute most to learning new tasks. Exploring these adaptation patterns across architectures may provide further insights into representation specialization and guide the design of future parameter-efficient adaptation methods.

\paragraph{Key Takeaways.}
Our work shows that residual-space adaptation extends naturally to stochastic diffusion models, enabling continual concept personalization with stable performance, negligible interference, and strong parameter efficiency. Beyond its empirical performance, the residual-space dynamics observed during continual adaptation reveal how adaptation demand is distributed across the adapted layers. By identifying the layers that consistently require new concept-specific adaptation directions, it provides a structural interpretation of existing adaptation strategies and directly guides more parameter-efficient adaptation. We hope this perspective motivates future work that uses adaptation dynamics not only to improve continual learning algorithms, but also to better understand layer specialization in modern generative models.

%% file: supp/1_metric_comparison.tex
\section{Embedding Metric Comparison}
\label{sup:metric_comparison}

We quantitatively compare the embedding spaces used to compute concept
fidelity in the main text. The goal of this comparison is to
measure whether an embedding model captures concept-level identity rather than
only low-level visual similarity. A suitable embedding should map images of the
same personalized concept close to one another while separating images from
different concepts.

We evaluate this property using a clustering protocol on the ground-truth
reference images. Given image embeddings, we define the pairwise clustering
distance as
\begin{equation}
d(i_1,i_2) = 1 - \mathrm{sim}(i_1,i_2),
\end{equation}
where $\mathrm{sim}(\cdot,\cdot)$ denotes the cosine similarity in the
corresponding embedding space. We cluster all ground-truth images into 101
clusters, matching the number of personalized concepts. To make the comparison
deterministic and reduce sensitivity to random initialization, the cluster
centroids are initialized using one image from each ground-truth concept.

We report clustering accuracy and Adjusted Rand Index (ARI) with respect to the
ground-truth concept labels. Clustering accuracy is computed as the fraction of
images assigned to their correct concept cluster. ARI measures the agreement between the predicted clustering and the ground-truth partition based on pairwise co-clustering decisions, adjusted for chance. Higher values indicate that the embedding space better recovers the underlying concept structure.

\begin{table}[!htbp]
\centering
\begin{tabular}{lcc}
\hline
Embedding model & Acc. & ARI \\
\hline
DINOv3 ViT-S/16 & 0.9085 & 0.8304 \\
DINO ViT-B/16 & 0.9196 & 0.8044 \\
OpenCLIP ViT-B/32 & 0.9543 & 0.8935 \\
CLIP ViT-B/32 & 0.9385 & 0.8873 \\
DreamSim ensemble & \textbf{0.9558} & \textbf{0.9050} \\
\hline
\end{tabular}
\caption{Clustering-based comparison of embedding spaces on ground-truth
reference images.}
\label{tab:embedding_metric_comparison}
\end{table}
As shown in Table~\ref{tab:embedding_metric_comparison}, DreamSim achieves the
highest performance on both metrics. This quantitative result complements the
nearest-neighbor visualization in the main text and supports the use of
DreamSim as the default embedding for concept fidelity.

%% file: supp/2_stochastic_derivation.tex
\section{Stochastic Residual-Space Derivation}
\label{sup:stochastic_derivation}

This section derives the residual-space construction used in the main text as
the solution of a stochastic adaptation objective. We focus on one adapted layer
$l$ and omit the layer superscript when the notation is unambiguous.

\subsection{Stochastic Objective}
\label{sup:stochastic_objective}

For concept $c_t$, the layer input in a diffusion model depends on the sampled
training image, the diffusion timestep $s$, and the Gaussian noise $\epsilon$.
We therefore treat the layer input as a random variable
\begin{equation}
    x_t^{(l)} = x^{(l)}(c_t,s,\epsilon),
    \qquad
    \epsilon \sim \mathcal N(0,I),
\end{equation}
which induces a concept-level feature distribution $p_t^{(l)}(x)$. Let
\begin{equation}
    M_t^{(l)}
    =
    \mathbb E_{x_t \sim p_t^{(l)}}[x_t x_t^\top]
\end{equation}
denote its second-moment matrix.

The new projection $B_t^{(l)}$ should satisfy two competing objectives. For the
current concept, it should preserve as much projected feature energy as
possible:
\begin{equation}
\max_{B_t^{(l)}}
\mathbb E_{x_t \sim p_t^{(l)}}
\left[
\|B_t^{(l)}x_t\|^2
\right],
\qquad
B_t^{(l)}(B_t^{(l)})^\top=I.
\label{eq:sup_max_adapt}
\end{equation}
For a previous concept $c_i$, $i<t$, it should instead have small expected
response:
\begin{equation}
\min_{B_t^{(l)}}
\mathbb E_{x_i \sim p_i^{(l)}}
\left[
\|B_t^{(l)}x_i\|^2
\right],
\qquad
B_t^{(l)}(B_t^{(l)})^\top=I.
\label{eq:sup_min_interference}
\end{equation}
For any concept $c_i$, this expected response can be written as
\begin{align}
\mathbb E_{x_i\sim p_i^{(l)}}
\left[
\|B_t^{(l)}x_i\|^2
\right]
&=
\mathbb E
\left[
x_i^\top (B_t^{(l)})^\top B_t^{(l)} x_i
\right] \nonumber \\
&=
\mathrm{Tr}
\left(
B_t^{(l)}M_i^{(l)}(B_t^{(l)})^\top
\right).
\label{eq:sup_expected_response_trace}
\end{align}

Let the eigenvalues of $M_i^{(l)}$ be sorted in descending order. By the
Rayleigh--Ritz variational principle, the exact rank-$r$ minimizer of
Eq.~\eqref{eq:sup_min_interference} is the bottom-$r$ eigenspace of
$M_i^{(l)}$, while the exact rank-$r$ maximizer of Eq.~\eqref{eq:sup_max_adapt}
is the top-$r$ eigenspace of $M_t^{(l)}$. Enforcing the bottom eigenspace
solution simultaneously for all previous concepts is impractical. DyRA instead
uses a tractable relaxation: after a task selects its dominant adaptation
directions, later tasks are constrained to be orthogonal to those directions.
This suppresses the directions with largest expected response on previous
concepts while leaving a residual subspace for new concepts.

\subsection{Residual-Space Construction}
\label{sup:residual_space_solution}

For the first task, there is no previous occupied subspace. The method therefore
solves only the adaptability objective in Eq.~\eqref{eq:sup_max_adapt}. Its
rank-$r_1^{(l)}$ solution is
\begin{equation}
    \mathcal R_1^{(l)}
    =
    \operatorname{TopEig}_{r_1^{(l)}}
    \left(
    M_1^{(l)}
    \right),
    \qquad
    \mathrm{Row}(B_1^{(l)})=\mathcal R_1^{(l)}.
    \label{eq:sup_task_one_solution}
\end{equation}
We set $\mathcal M_1^{(l)}=\mathcal R_1^{(l)}$.

For task $t>1$, let $\mathcal M_{t-1}^{(l)}$ denote the accumulated occupied
subspace from previous tasks and let $P_{t-1}^{(l)}$ be its orthogonal
projector. The relaxed interference objective constrains the new projection to
avoid previously selected dominant directions:
\begin{equation}
    \mathrm{Row}(B_t^{(l)})
    \subseteq
    \left(\mathcal M_{t-1}^{(l)}\right)^\perp .
    \label{eq:sup_relaxed_constraint}
\end{equation}
This constraint implies
\begin{equation}
    B_t^{(l)}P_{t-1}^{(l)} = 0.
\end{equation}
Therefore the current feature can be replaced by its residual component in the
adaptability objective:
\begin{equation}
    B_t^{(l)}x_t
    =
    B_t^{(l)}
    (I-P_{t-1}^{(l)})x_t .
\end{equation}
The constrained current-task objective becomes
\begin{equation}
\begin{aligned}
\max_{B_t^{(l)}} \quad &
\mathbb E
\left[
\left\|
B_t^{(l)}
(I-P_{t-1}^{(l)})x_t
\right\|^2
\right] \\
\text{s.t.}\quad &
\mathrm{Row}(B_t^{(l)})
\subseteq
\left(\mathcal M_{t-1}^{(l)}\right)^\perp,\\
&
B_t^{(l)}(B_t^{(l)})^\top=I .
\end{aligned}
\label{eq:sup_residual_energy_objective}
\end{equation}
The corresponding residual second-moment matrix is
\begin{equation}
    M_{t,\perp}^{(l)}
    =
    (I-P_{t-1}^{(l)})
    M_t^{(l)}
    (I-P_{t-1}^{(l)}).
    \label{eq:sup_residual_second_moment}
\end{equation}
Using Eq.~\eqref{eq:sup_expected_response_trace}, the objective in
Eq.~\eqref{eq:sup_residual_energy_objective} becomes
\begin{equation}
\begin{aligned}
\max_{B_t^{(l)}} \quad &
\mathrm{Tr}
\left(
B_t^{(l)}
M_{t,\perp}^{(l)}
(B_t^{(l)})^\top
\right) \\
\text{s.t.}\quad &
\mathrm{Row}(B_t^{(l)})
\subseteq
\left(\mathcal M_{t-1}^{(l)}\right)^\perp,\\
&
B_t^{(l)}(B_t^{(l)})^\top=I .
\end{aligned}
\label{eq:sup_trace_max}
\end{equation}
By the Rayleigh--Ritz variational principle, the optimal rank-$r_t^{(l)}$ row
space is the dominant eigenspace of the residual second-moment matrix:
\begin{equation}
    \mathcal R_t^{(l)}
    =
    \operatorname{TopEig}_{r_t^{(l)}}
    \left(
    M_{t,\perp}^{(l)}
    \right),
    \qquad
    \mathrm{Row}(B_t^{(l)})=\mathcal R_t^{(l)}.
    \label{eq:sup_residual_solution}
\end{equation}
The occupied subspace is then updated recursively as
\begin{equation}
    \mathcal M_t^{(l)}
    =
    \mathcal M_{t-1}^{(l)}
    \oplus
    \mathcal R_t^{(l)}.
\end{equation}

Consequently, the DyRA construction is the natural solution of the stochastic
max/min adaptation problem under the practical interference relaxation: task 1
selects the dominant eigenspace of its stochastic features, and each later task
selects the dominant eigenspace of the current stochastic features after
projecting out directions occupied by previous tasks.

\subsection{Empirical Estimation and Initialization}
\label{sup:dyra_algorithm}

In implementation, $M_t^{(l)}$ is estimated from collected stochastic layer
features. If $H_t^{(l)}=[x_{t,1}^{(l)},\dots,x_{t,n}^{(l)}]$ is the feature
matrix collected before training task $t$, we use
\begin{equation}
    \widehat M_t^{(l)}
    =
    \frac{1}{n}H_t^{(l)}(H_t^{(l)})^\top .
\end{equation}
The samples used to construct $H_t^{(l)}$ follow the same stochastic process as
diffusion training: each forward pass uses sampled Gaussian noise and a sampled
denoising timestep. Therefore $\widehat M_t^{(l)}$ is an empirical estimate of
the second moment appearing in the stochastic objective.

Algorithm~\ref{alg:sup_dyra_init} summarizes the initialization. The rank is
selected independently for each layer by thresholding the singular values of
the residual second-moment estimate, so a single threshold $\tau$ induces
layer-dependent ranks $r_t^{(l)}$.

\begin{algorithm}[!t]
\SetAlgoLined
\KwIn{adapted layers $\mathcal L_{\mathrm{adapt}}$; current data
$\mathcal D_t$; threshold $\tau$; task index $t$; layer bases
$Q^{(l)}$ and allocated ranks $\mathrm{cumRank}^{(l)}$, initialized as
empty and zero before task 1}
\KwOut{fixed LoRA projections $B_t^{(l)}$; updated bases $Q^{(l)}$ and ranks
$\mathrm{cumRank}^{(l)}$}

Collect stochastic layer features by forwarding $\mathcal D_t$ with sampled
diffusion timesteps and Gaussian noise\;
Compute $\widehat M_t^{(l)} = \frac{1}{n}H_t^{(l)}(H_t^{(l)})^\top$
for each $l\in\mathcal L_{\mathrm{adapt}}$\;

\ForEach{$l\in\mathcal L_{\mathrm{adapt}}$}{
    \eIf{$t=1$}{
        $C^{(l)} \leftarrow \widehat M_t^{(l)}$\;
    }{
        \lIf{$\mathrm{cumRank}^{(l)} \ge d_{\mathrm{in}}^{(l)}$}{
        \textbf{continue}}
        $P^{(l)} \leftarrow Q^{(l)}(Q^{(l)})^\top$\;
        $C^{(l)} \leftarrow (I-P^{(l)})\widehat M_t^{(l)}(I-P^{(l)})$\;
    }

    $U,S \leftarrow \mathrm{SVD}(C^{(l)})$\;
    $r \leftarrow |\{k:S_k\ge\tau\}|$\;
    \eIf{$t=1$}{
        \lIf{$r=0$}{$r\leftarrow 1$}
    }{
        \lIf{$r=0$}{\textbf{continue}}
        $r \leftarrow \min(r,\, d_{\mathrm{in}}^{(l)}-\mathrm{cumRank}^{(l)})$\;
    }

    $U_r \leftarrow U[:,1:r]$\;
    $B_t^{(l)} \leftarrow U_r^\top$\;
    Initialize $A_t^{(l)}\leftarrow 0$\;
    $Q^{(l)} \leftarrow \mathrm{QR}([Q^{(l)}\,|\,U_r])$\;
    $\mathrm{cumRank}^{(l)} \leftarrow \mathrm{cumRank}^{(l)}+r$\;
}
\caption{DyRA initialization for stochastic diffusion features.}
\label{alg:sup_dyra_init}
\end{algorithm}
We maintain $Q^{(l)}$ as an orthonormal basis of the occupied subspace and form
the projector as $P^{(l)}=Q^{(l)}(Q^{(l)})^\top$. This avoids accumulating
numerical error from previously initialized LoRA matrices and keeps the
residual covariance in Eq.~\eqref{eq:sup_residual_second_moment} symmetric
positive semidefinite up to numerical precision.

%% file: supp/3_implementation_details.tex
\section{Implementation Details}
\label{sup:implementation_details}

We provide additional implementation details for the continual personalization
experiments in the main text.

\paragraph{Dataset and task order.}
We randomly shuffle the 101 concepts once and use this fixed task order for all
methods and random seeds in the main experiments. This avoids presenting
concepts in semantic clusters, such as all visually similar instances from the
same class arriving consecutively. During each task, the dataloader contains
only the reference images for the current concept. To check robustness to task
ordering, we additionally repeat all experiments with two other independently
shuffled concept orders and observe results consistent with those reported in
the main text.

\paragraph{Backbone and adapted layers.}
We use Stable Diffusion v1.5~\cite{rombach2022high} as the pretrained
text-to-image backbone. The model uses a CLIP ViT-L/14 text encoder with 12
transformer blocks and a UNet with 16 cross-attention blocks. LoRA modules are
inserted into the query, key, value, and output projections of the attention
layers in both the text encoder and the UNet. This yields 48 adapted projection
layers in the text encoder and 128 adapted projection layers in the UNet.

\paragraph{Training configuration.}
Each concept is trained for 400 gradient-update steps. Since each concept has
only a small number of reference images, the concept dataset is cycled during
training. We train with fp16 mixed precision on two NVIDIA RTX 4090 GPUs. All
methods are run with three global random seeds, $\{0,42,619\}$. The training
prompt for concept $c$ follows the dual-token format
\begin{equation}
\text{``a photo of } \langle c_1\rangle \langle c_2\rangle \text{''}.
\end{equation}

\begin{table}[!htbp]
\centering
\small
\begin{tabular}{p{0.43\linewidth}p{0.47\linewidth}}
\hline
Setting & Value \\
\hline
Training steps per task & 400 \\
Evaluation interval & Every 50 steps \\
Per-device batch size & 2 \\
Optimizer & AdamW \\
AdamW parameters & $\beta_1{=}0.9$, $\beta_2{=}0.999$, wd $0.01$ \\
\multicolumn{2}{l}{\textit{Learning rate}} \\
LR schedule & Linear decay, no warmup \\
Token-embedding LR & $1\times10^{-3}$ \\
Text-encoder LoRA LR & $1\times10^{-5}$ \\
UNet LoRA LR & $1\times10^{-4}$ \\
\multicolumn{2}{l}{\textit{LoRA configuration}} \\
Rank (fixed-rank baselines) & $r=4$ \\
Scaling coefficient & $\alpha=1.0$ \\
\multicolumn{2}{l}{\textit{Image preprocessing}} \\
Input resolution & $512\times512$ \\
Random crop & $p{=}0.5$ \\
Random scale & $0.8$--$1.0$ \\
\hline
\end{tabular}
\caption{Training hyperparameters shared across methods.}
\label{tab:sup_training_config}
\end{table}
\paragraph{Residual-space construction.}
Layer inputs for subspace estimation are collected before training each new
task using the same stochastic inputs as diffusion training: Gaussian noise is
sampled for each image and the denoising timestep is sampled uniformly.

\paragraph{Evaluation protocol.}
We evaluate the model every 50 training steps. At each evaluation checkpoint,
we generate images for all concepts that have been learned so far. For each
concept, we sample 10 images using different random seeds. The generated images
are compared with the ground-truth reference images using the concept-fidelity
metric described in the main text. During this quantitative evaluation, we
prompt the model only with the two concept-specific tokens,
$\langle c_1\rangle\langle c_2\rangle$, to isolate concept identity from
prompt-induced scene variation.

\paragraph{Final qualitative validation.}
For final qualitative validation after all 101 concepts have been learned, we
use the class-level prompt templates provided by Concept 101. Each semantic
class has 10 prompts, and we replace the placeholder token in each template
with the two concept-specific tokens. We sample 8 images per prompt with
different random seeds. For the \textit{pet\_dog2} validation example shown in
the main text, the five selected prompts are \texttt{<TOK> at a beach with a
view of the seashore.}, \texttt{<TOK> in Times Square.}, \texttt{<TOK> is in a
construction outfit.}, \texttt{<TOK> is playing with a ball.}, and
\texttt{<TOK> is wearing a sombrero.} In all templates, \texttt{<TOK>} is
replaced by the two concept-specific tokens. All methods use the same prompts
and seeds.

\begin{table}[!htbp]
\centering
\small
\begin{tabular}{p{0.43\linewidth}p{0.47\linewidth}}
\hline
Setting & Value \\
\hline
Sampler & DPM-Solver++ multistep \\
Inference steps & 50 \\
CFG scale & 7.5 \\
Output resolution & $512\times512$ \\
Evaluation samples per concept & 10 \\
Validation prompts per class & 10 \\
Validation samples per prompt & 8 \\
\hline
\end{tabular}
\caption{Generation settings used for evaluation and final validation.}
\label{tab:sup_inference_config}
\end{table}

%% file: supp/4_layer_mapping.tex
\section{Layer-Index Mapping}
\label{sup:layer_mapping}

This section provides the mapping between the adapted layer indices used in the
layer-wise rank analysis and the corresponding attention projections. We use
$Q$, $K$, $V$, and $O$ to denote the query, key, value, and output projections,
respectively. For the UNet, we explicitly distinguish self-attention from
cross-attention projections.

\paragraph{Text encoder.}
The CLIP text encoder contains 12 transformer blocks. In each block, we adapt
the self-attention projections in the order $K,V,Q,O$. Thus, for text-encoder
block $b\in\{0,\dots,11\}$, the adapted layer indices are
\begin{equation}
4b + \{0,1,2,3\}
\quad\longleftrightarrow\quad
\{K,V,Q,O\}.
\end{equation}
All text-encoder adapted layers have input dimension $d_{\mathrm{in}}=768$.

\begin{table}[!htbp]
\centering
\small
\begin{tabular}{c c c c c c}
\hline
Block $b$ & $K$ & $V$ & $Q$ & $O$ & $d_{\mathrm{in}}$ \\
\hline
0  & 0  & 1  & 2  & 3  & 768 \\
1  & 4  & 5  & 6  & 7  & 768 \\
2  & 8  & 9  & 10 & 11 & 768 \\
3  & 12 & 13 & 14 & 15 & 768 \\
4  & 16 & 17 & 18 & 19 & 768 \\
5  & 20 & 21 & 22 & 23 & 768 \\
6  & 24 & 25 & 26 & 27 & 768 \\
7  & 28 & 29 & 30 & 31 & 768 \\
8  & 32 & 33 & 34 & 35 & 768 \\
9  & 36 & 37 & 38 & 39 & 768 \\
10 & 40 & 41 & 42 & 43 & 768 \\
11 & 44 & 45 & 46 & 47 & 768 \\
\hline
\end{tabular}
\caption{Text-encoder layer-index mapping. The text encoder uses
self-attention only.}
\label{tab:sup_text_encoder_mapping}
\end{table}
\paragraph{UNet.}
The UNet contains 16 adapted transformer modules across the down, up, and
middle blocks. Each module contributes eight adapted projection layers: four
self-attention projections followed by four cross-attention projections. For
UNet transformer module $m$, the self-attention indices are
\begin{equation}
8m+\{0,1,2,3\}
\longleftrightarrow
\{\mathrm{self}\text{-}Q,\mathrm{self}\text{-}K,
\mathrm{self}\text{-}V,\mathrm{self}\text{-}O\},
\end{equation}
and the cross-attention indices are
\begin{equation}
\begin{gathered}
8m+\{4,5,6,7\}
\longleftrightarrow
\\
\{\mathrm{cross}\text{-}Q,\mathrm{cross}\text{-}K,
\mathrm{cross}\text{-}V,\mathrm{cross}\text{-}O\}.
\end{gathered}
\end{equation}
The cross-attention $K$ and $V$ projections take text-encoder features as
input, so their input dimension is always $768$. The remaining projections use
the UNet hidden dimension of the corresponding transformer module.

\begin{table}[!htbp]
\centering
\footnotesize
\setlength{\tabcolsep}{2pt}
\begin{tabular}{c c c c}
\hline
$m$ & Range & Block & UNet dim. \\
\hline
0  & 0--7     & down 0 & 320  \\
1  & 8--15    & down 0 & 320  \\
2  & 16--23   & down 1 & 640  \\
3  & 24--31   & down 1 & 640  \\
4  & 32--39   & down 2 & 1280 \\
5  & 40--47   & down 2 & 1280 \\
6  & 48--55   & up 1   & 1280 \\
7  & 56--63   & up 1   & 1280 \\
8  & 64--71   & up 1   & 1280 \\
9  & 72--79   & up 2   & 1280 \\
10 & 80--87   & up 2   & 1280 \\
11 & 88--95   & up 2   & 1280 \\
12 & 96--103  & up 3   & 320  \\
13 & 104--111 & up 3   & 320  \\
14 & 112--119 & up 3   & 320  \\
15 & 120--127 & middle & 1280 \\
\hline
\end{tabular}
\caption{UNet transformer-module mapping. Within each layer range, offsets
$0$--$3$ correspond to self-attention $Q,K,V,O$, and offsets $4$--$7$
correspond to cross-attention $Q,K,V,O$.}
\label{tab:sup_unet_mapping}
\end{table}
This mapping explains the periodic structure in the rank-allocation plots. In
particular, the UNet cross-attention $K,V$ projections correspond to indices
$8m+5$ and $8m+6$ for each transformer module $m$, which are the layers that
consistently receive the largest residual ranks in the analysis.

%% file: supp/5_qualitative_examples.tex
\section{Additional Qualitative Examples}
\label{sup:qualitative_examples}

We provide additional final-validation and continual-generation examples for
three other concepts spanning different categories. For each concept, the
continual-generation figure is shown first, followed by the final validation
generations. In the validation figures, columns $P_1$--$P_5$ correspond to the
prompt set listed in the same subsection, and \texttt{<TOK>} is replaced by the
two learned concept tokens.
For each continual-generation figure, the first shown task is the task at
which the concept is introduced.

\clearpage
\subsection{Jewelry Ring}
\label{sup:qual_jewelryring}

\paragraph{Validation prompts.}
\begin{enumerate}[leftmargin=*,label=$P_{\arabic*}$:,itemsep=1pt,topsep=2pt]
    \item A close shot of \texttt{<TOK>} worn by a model.
    \item A \texttt{<TOK>} lying in a red jewelry box.
    \item A \texttt{<TOK>} on a blue velvet rug.
    \item A \texttt{<TOK>} on a mossy forest floor.
    \item A \texttt{<TOK>} on a sunlit beach, a seashell beside it.
\end{enumerate}

\begin{figure}[H]
\centering
\begin{subfigure}{0.52\linewidth}
\centering
\includegraphics[width=\linewidth,clip]{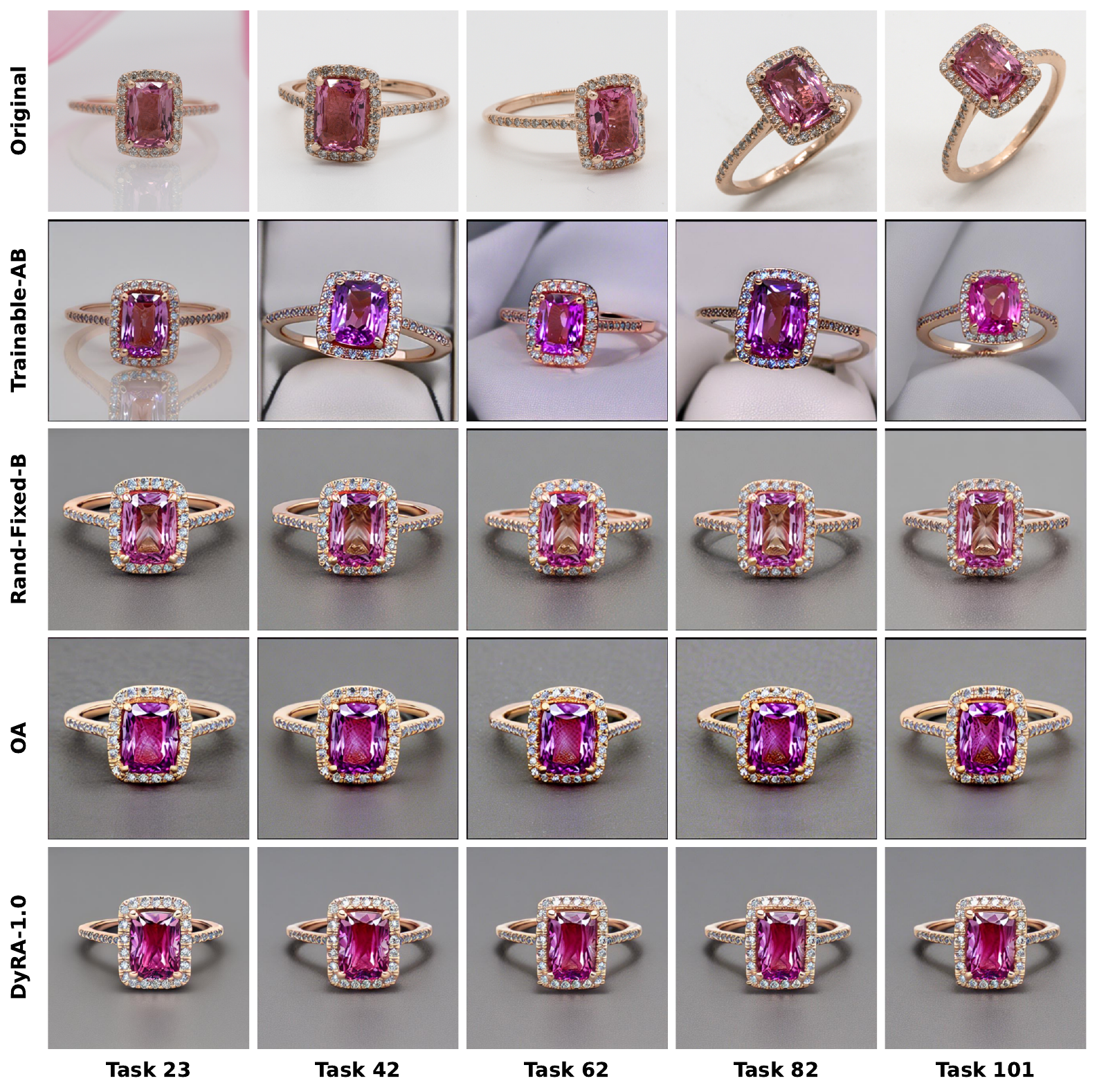}
\caption{Continual generation, beginning at task~23.}
\label{fig:sup_continual_jewelryring}
\end{subfigure}
\par\medskip
\begin{subfigure}{0.52\linewidth}
\centering
\includegraphics[width=\linewidth,clip]{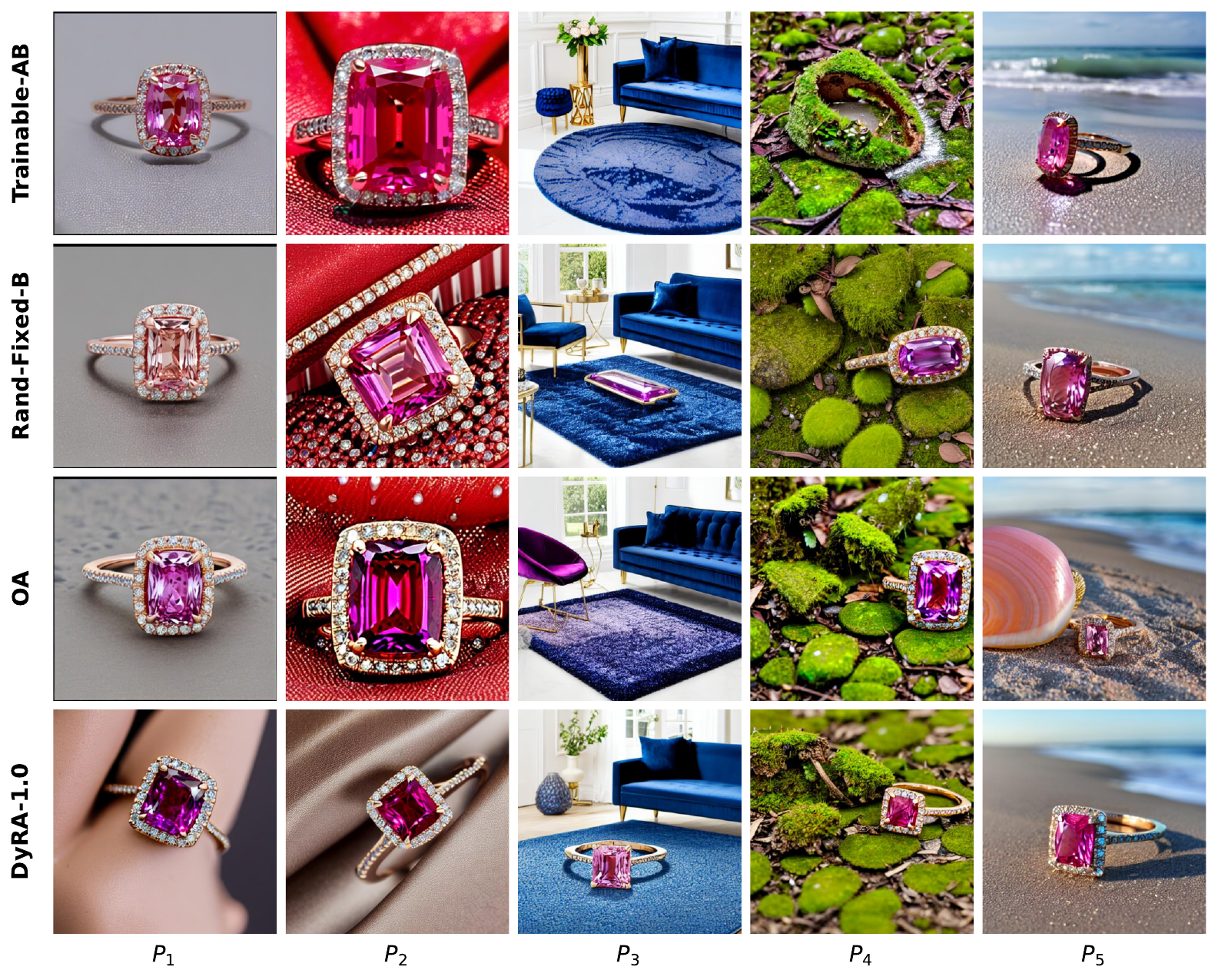}
\caption{Final validation; columns correspond to prompts $P_1$--$P_5$.}
\label{fig:sup_validation_jewelryring}
\end{subfigure}
\caption{Additional qualitative results for Jewelry Ring.}
\end{figure}

\clearpage
\subsection{Plushie Panda}
\label{sup:qual_plushiepanda}

\noindent\textbf{Validation prompts.}\par
\smallskip
\noindent\begin{tabular}{@{}r@{ }l@{}}
$P_1$: & Photo of a \texttt{<TOK>}. \\
$P_2$: & \texttt{<TOK>} in front of a medieval castle. \\
$P_3$: & \texttt{<TOK>} in Grand Canyon. \\
$P_4$: & \texttt{<TOK>} in Times Square. \\
$P_5$: & \texttt{<TOK>} on a boat in the sea.
\end{tabular}

\begin{figure}[H]
\centering
\begin{subfigure}{0.52\linewidth}
\centering
\includegraphics[width=\linewidth,clip]{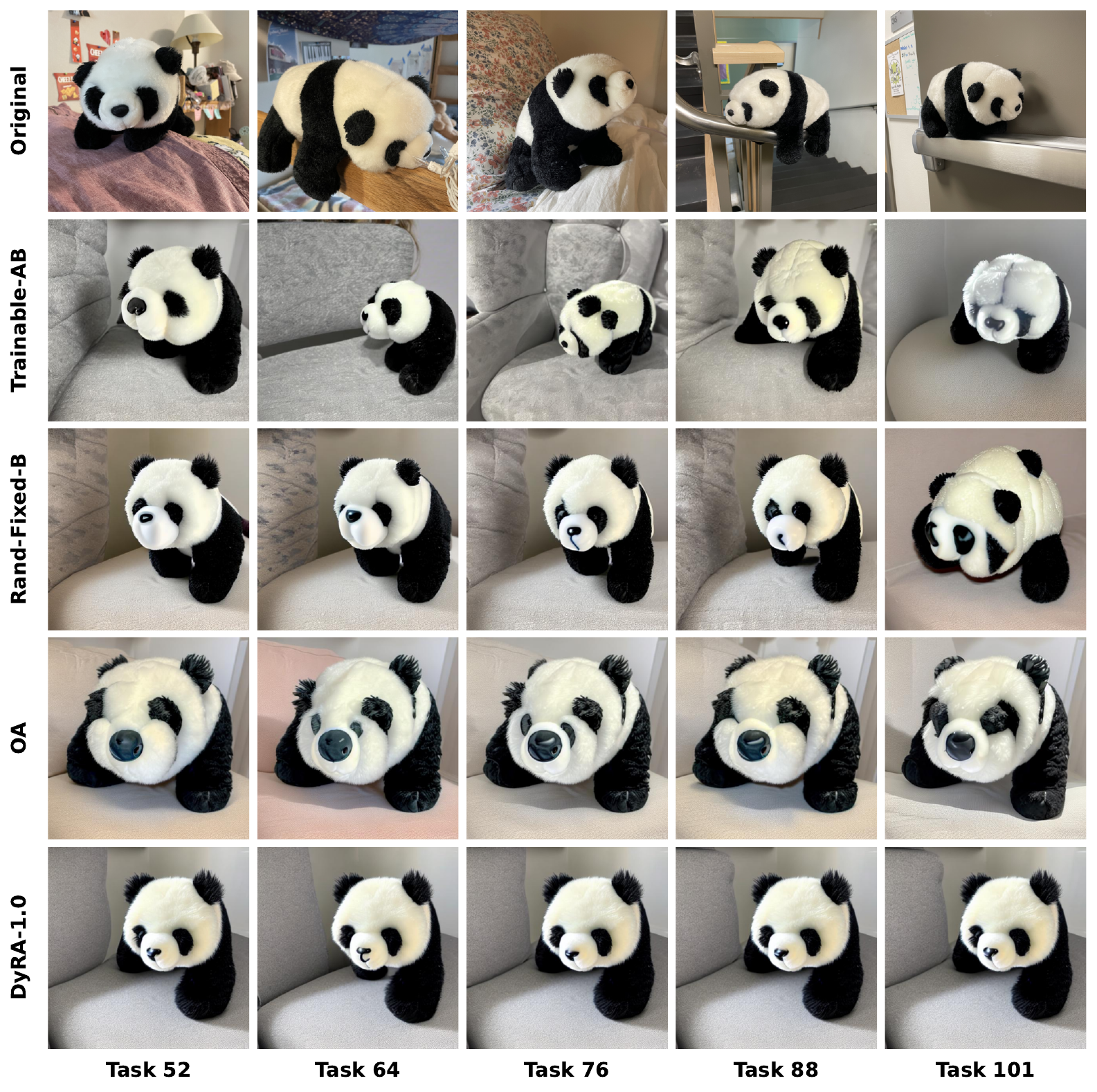}
\caption{Continual generation, beginning at task~52.}
\label{fig:sup_continual_plushiepanda}
\end{subfigure}
\par\medskip
\begin{subfigure}{0.52\linewidth}
\centering
\includegraphics[width=\linewidth,clip]{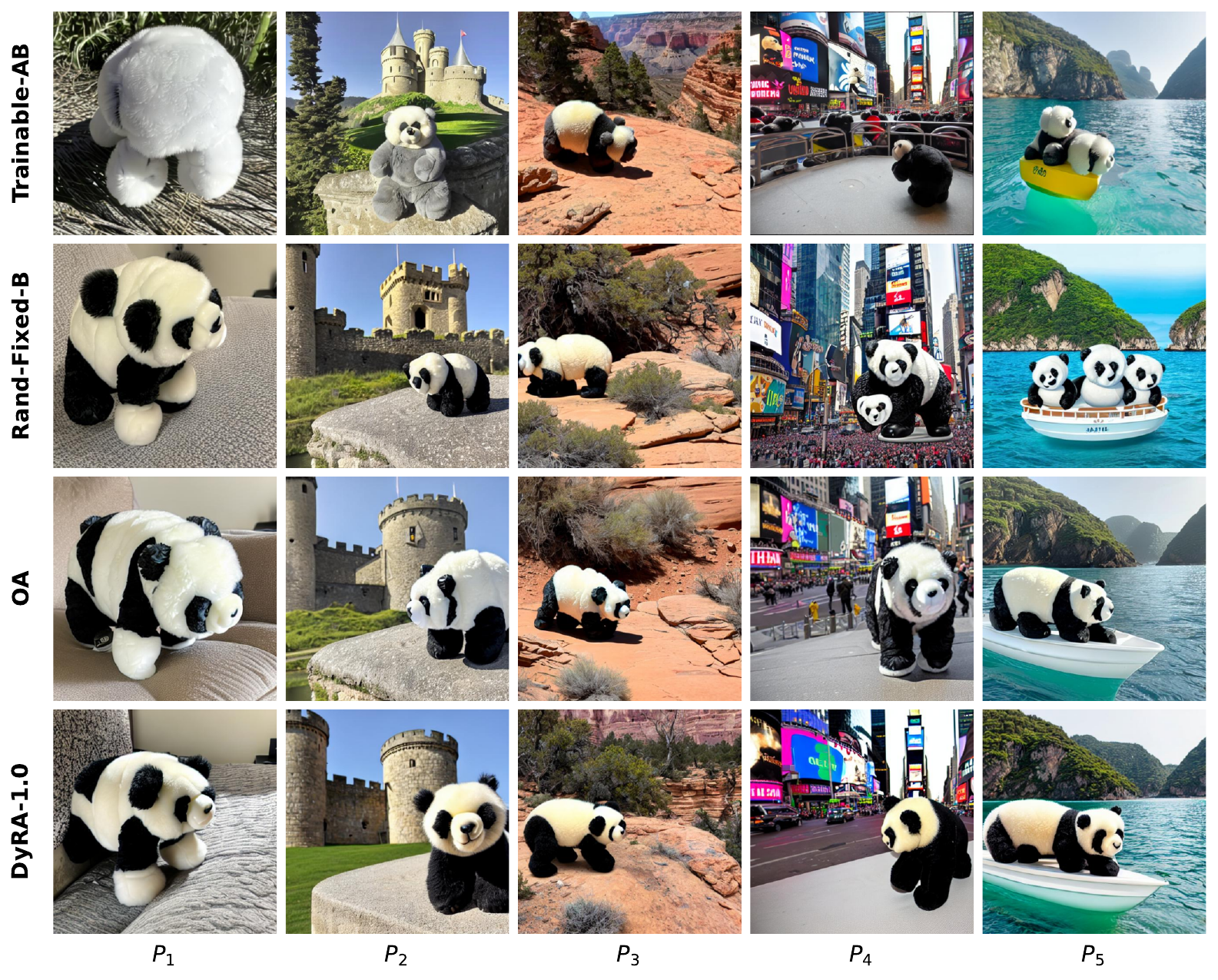}
\caption{Final validation; columns correspond to prompts $P_1$--$P_5$.}
\label{fig:sup_validation_plushiepanda}
\end{subfigure}
\caption{Additional qualitative results for Plushie Panda.}
\end{figure}

\clearpage
\subsection{Wearable Jacket}
\label{sup:qual_wearablejacket1}

\paragraph{Validation prompts.}
\begin{enumerate}[leftmargin=*,label=$P_{\arabic*}$:,itemsep=1pt,topsep=2pt]
    \item A humanoid robot wearing \texttt{<TOK>} with a futuristic city in the
    background.
    \item A mannequin showcasing \texttt{<TOK>}.
    \item A \texttt{<TOK>} on a hanger in the window of a shop.
    \item A \texttt{<TOK>} on a hook in a study room with bookshelves.
    \item Photo of a \texttt{<TOK>}.
\end{enumerate}

\begin{figure}[H]
\centering
\begin{subfigure}{0.52\linewidth}
\centering
\includegraphics[width=\linewidth,clip]{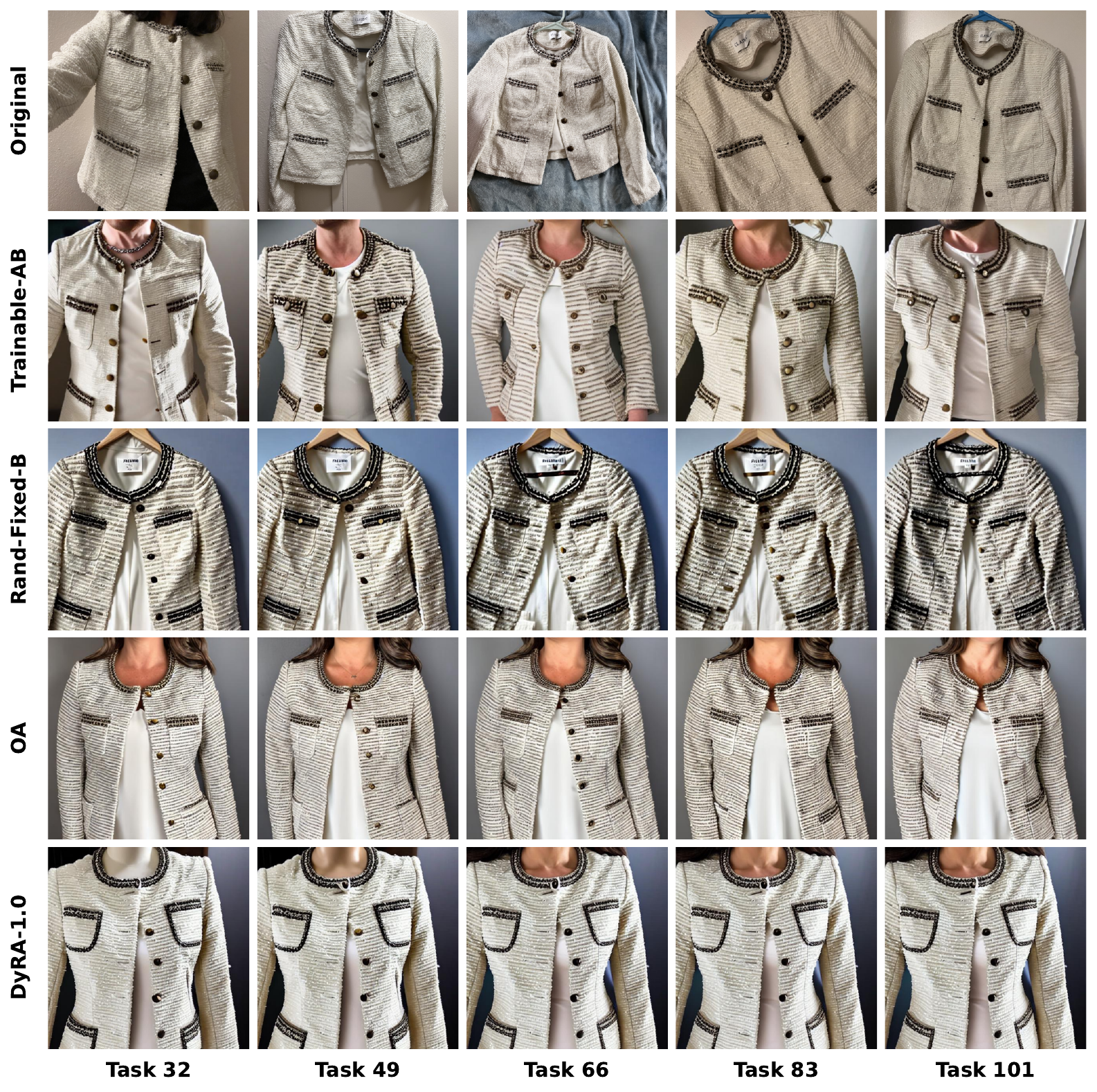}
\caption{Continual generation, beginning at task~32.}
\label{fig:sup_continual_wearablejacket1}
\end{subfigure}
\par\medskip
\begin{subfigure}{0.52\linewidth}
\centering
\includegraphics[width=\linewidth,clip]{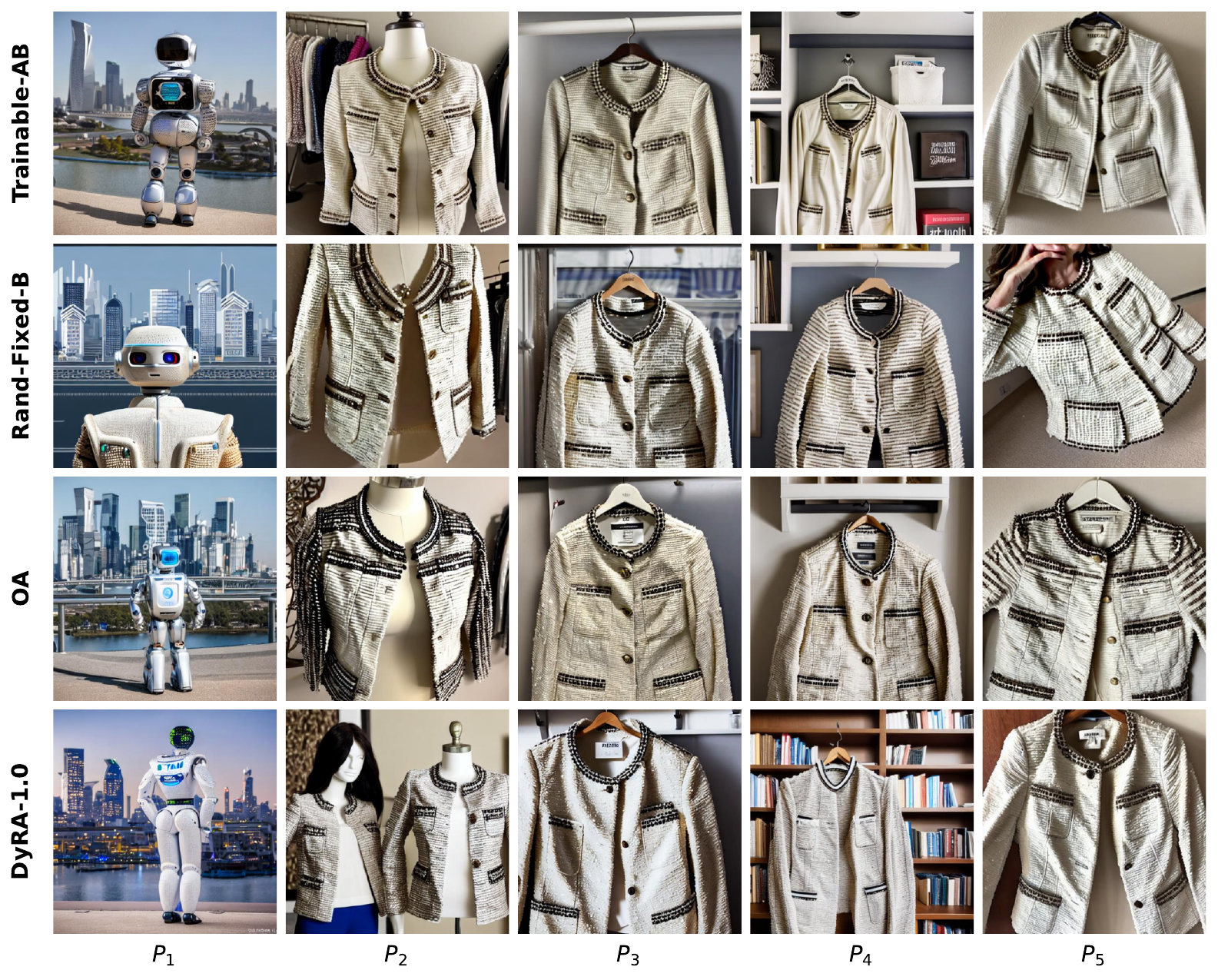}
\caption{Final validation; columns correspond to prompts $P_1$--$P_5$.}
\label{fig:sup_validation_wearablejacket1}
\end{subfigure}
\caption{Additional qualitative results for Wearable Jacket.}
\end{figure}
\FloatBarrier